\documentclass[runningheads]{llncs}
\usepackage{graphicx}
\usepackage{tikz}
\usepackage{comment}
\usepackage{amsmath,amssymb} % define this before the line numbering.
\usepackage{color}
\usepackage{algorithm}
\usepackage{algpseudocode}
\usepackage{url}
\usepackage{subcaption}
\usepackage{bbm}
\usepackage{setspace}
\usepackage{float}
\usepackage{hyperref}
\usepackage{orcidlink}

\usepackage[accsupp]{axessibility}  % Improves PDF readability for those with disabilities.

\usepackage[capitalize]{cleveref}
\crefname{section}{Sec.}{Secs.}
\Crefname{section}{Section}{Sections}
\Crefname{table}{Table}{Tables}
\crefname{table}{Tab.}{Tabs.}

\def\eg{\emph{e.g.}} 
\def\ie{\emph{i.e.}} 
 
\def\etc{\emph{etc.}} \def\vs{\emph{vs.}}

\def\etal{\emph{et al.}}

\newcommand{\footlabel}[2]{%
    \addtocounter{footnote}{1}%
    \footnotetext[\thefootnote]{%
        \addtocounter{footnote}{-1}%
        \refstepcounter{footnote}\label{#1}%
        #2%
    }%
    $^{\ref{#1}}$%
}

\newcommand{\footref}[1]{%
    $^{\ref{#1}}$%
}

\begin{document}
% \renewcommand\thelinenumber{\color[rgb]{0.2,0.5,0.8}\normalfont\sffamily\scriptsize\arabic{linenumber}\color[rgb]{0,0,0}}
% \renewcommand\makeLineNumber {\hss\thelinenumber\ \hspace{6mm} \rlap{\hskip\textwidth\ \hspace{6.5mm}\thelinenumber}}
% \linenumbers
\pagestyle{headings}
\mainmatter
\def\ECCVSubNumber{4557}  % Insert your submission number here

\title{Adaptive Image Transformations for Transfer-based Adversarial Attack} % Replace with your title

% INITIAL SUBMISSION 
\begin{comment}
\titlerunning{ECCV-22 submission ID \ECCVSubNumber} 
\authorrunning{ECCV-22 submission ID \ECCVSubNumber} 
\author{Anonymous ECCV submission}
\institute{Paper ID \ECCVSubNumber}
\end{comment}
%******************

% CAMERA READY SUBMISSION
% \begin{comment}
\titlerunning{Adaptive Image Transformations for Transfer-based Adversarial Attack}
% If the paper title is too long for the running head, you can set
% an abbreviated paper title here
%
\author{Zheng Yuan\inst{1,2}\orcidlink{0000-0001-8788-6817} \and
Jie Zhang\inst{1,2,3}\orcidlink{0000-0002-8899-3996} \and
Shiguang Shan\inst{1,2}\orcidlink{0000-0002-8348-392X}}
\authorrunning{Z. Yuan et al.}
% First names are abbreviated in the running head.
% If there are more than two authors, 'et al.' is used.
%
\institute{Institute of Computing Technology, Chinese Academy of Sciences, Beijing, China \and University of Chinese Academy of Sciences, Beijing, China \and Institute of Intelligent Computing Technology, Suzhou, CAS, Suzhou, China
\email{zheng.yuan@vipl.ict.ac.cn} \quad \email{\{zhangjie,sgshan\}@ict.ac.cn}\\
}
% \end{comment}
%******************
\maketitle

\begin{abstract}
  Adversarial attacks provide a good way to study the robustness of deep learning models. One category of methods in transfer-based black-box attack utilizes several image transformation operations to improve the transferability of adversarial examples, which is effective, but fails to take the specific characteristic of the input image into consideration. In this work, we propose a novel architecture, called Adaptive Image Transformation Learner (AITL), which incorporates different image transformation operations into a unified framework to further improve the transferability of adversarial examples. Unlike the fixed combinational transformations used in existing works, our elaborately designed transformation learner adaptively selects the most effective combination of image transformations specific to the input image. Extensive experiments on ImageNet demonstrate that our method significantly improves the attack success rates on both normally trained models and defense models under various settings.
\end{abstract}

\section{Introduction}
\label{sec:intro}
  The field of deep neural networks has developed vigorously in recent years. The models have been successfully applied to various tasks, including image classification~\cite{he2016deep,szegedy2016rethinking,zoph2018learning}, face recognition~\cite{liu2017sphereface,wang2018cosface,deng2019arcface}, semantic segmentation~\cite{chen2015semantic,chen2018deeplab,chen2017rethinking}, \etc~However, the security of the DNN models raises great concerns due to that the model is vulnerable to adversarial examples~\cite{szegedy2014intriguing}. For example, an image with indistinguishable noise can mislead a well-trained classification model into the wrong category~\cite{goodfellow2015explaining}, or a stop sign on the road with a small elaborate patch can fool an autonomous vehicle~\cite{eykholt2018robust}. Adversarial attack and adversarial defense are like a spear and a shield. They promote the development of each other and together improve the robustness of deep neural networks.

\begin{table}
  \centering
  \caption{The list of image transformation methods used in various input-transformation-based adversarial attack methods}
  \resizebox{0.6\linewidth}{!}{
    \begin{tabular}{@{}cc|cc@{}}
      \hline
      \noalign{\smallskip}
      Method & Transformation & Method & Transformation \\
      \noalign{\smallskip}
      \hline
      \noalign{\smallskip}
      DIM~\cite{xie2019improving} & Resize & CIM~\cite{yang2021adversarial} & Crop \\
      TIM~\cite{dong2019evading} & Translate & Admix~\cite{wang2021admix} & Mixup \\
      SIM~\cite{lin2020nesterov} & Scale & AITL (ours) & Adaptive \\
      \noalign{\smallskip}
      \hline
    \end{tabular}
  }
  \label{tab:list}

  % \vspace{-7mm}
\end{table}

  Our work focuses on a popular scenario in the adversarial attack, \ie, transfer-based black-box attack. In this setting, the adversary can not get access to any information about the target model. Szegedy \etal~\cite{szegedy2014intriguing} find that adversarial examples have the property of cross model transferability, \ie, the adversarial example generated from a source model can also fool a target model. To further improve the transferability of adversarial examples, the subsequent works mainly adopt different input transformations~\cite{xie2019improving,dong2019evading,lin2020nesterov,wang2021admix} and modified gradient updates~\cite{dong2018boosting,lin2020nesterov,zou2020making,yang2021adversarial}. The former improves the transferability of adversarial examples by conducting various image transformations (\eg, resizing, crop, scale, mixup) on the original images before passing through the classifier. And the latter introduces the idea of various optimizers (\eg, momentum and NAG~\cite{sutskever2013on}, Adam~\cite{kingma2015adam}, AdaBelief~\cite{zhuang2020adabelief}) into the basic iterative attack method~\cite{kurakin2017adversarial} to improve the stability of the gradient and enhance the transferability of the generated adversarial examples.

\begin{figure*}[t]
  \centering
  \includegraphics[width=0.9\linewidth]{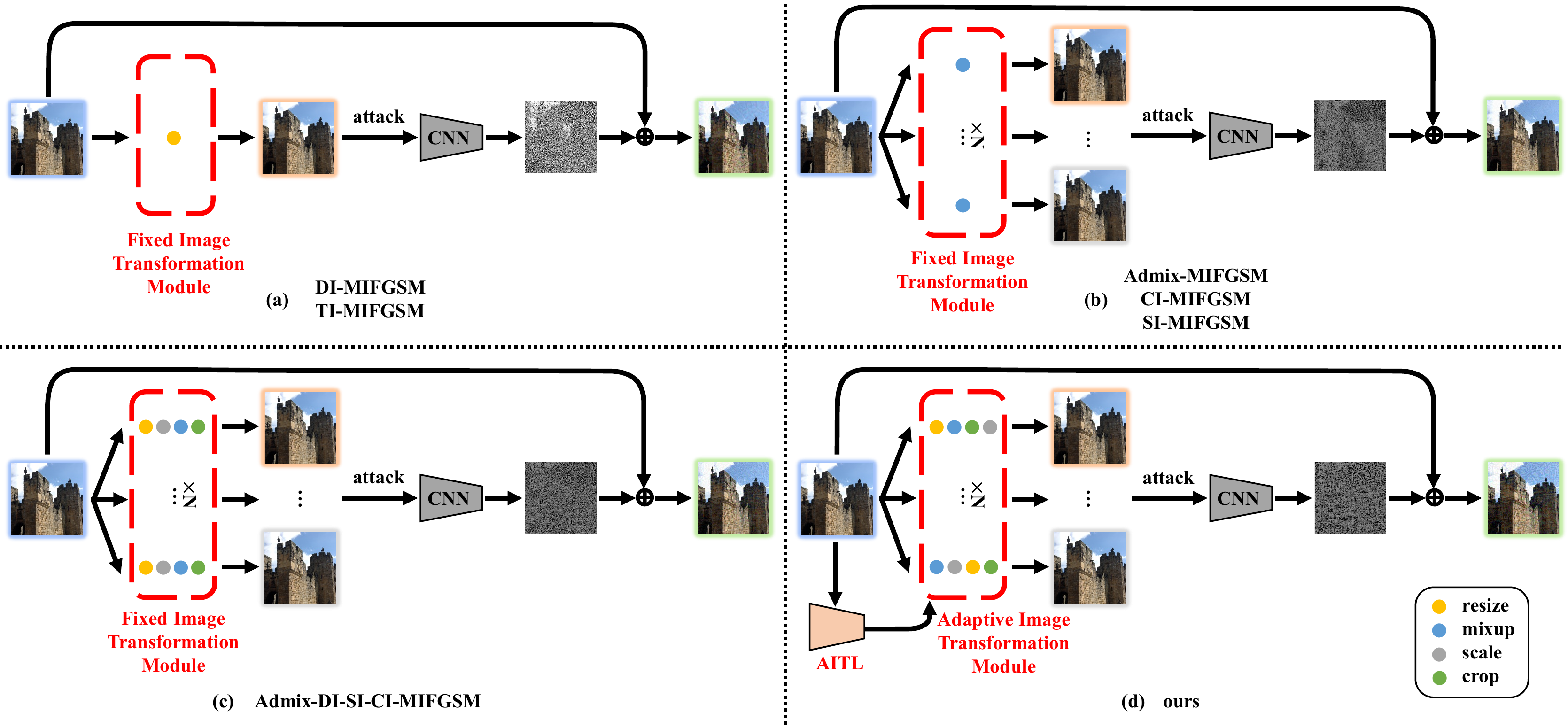}
  % \vspace{-8mm}
  \caption{Comparison between existing input-transformation-based black-box adversarial attack methods and our work. Different colors of the small circles in the red dotted box correspond to different image transformation operations. Existing works only conduct fixed image transformation once (as (a)) or repeat several times in parallel (as (b)), or simply combine multiple image transformation operations in the fixed sequence (as (c)). Our proposed method (as (d)) takes the characteristic of the current input image into consideration, utilizing an Adaptive Image Transformation Learner (AITL) to achieve the most effective combination of image transformations for each image, which can further improve the transferability of generated adversarial examples}
  \label{fig:comparison}
  % \vspace{-8mm}
\end{figure*}

  Existing transfer-based attack methods have studied a variety of image transformation operations, including resizing~\cite{xie2019improving}, crop~\cite{yang2021adversarial}, scale~\cite{lin2020nesterov} and so on (as listed in \cref{tab:list}). Although effective, we find that almost all existing works of input-transformation-based methods only investigate the effectiveness of fixed image transformation operations respectively (see (a) and (b) in \cref{fig:comparison}), or simply combine them in sequence (see (c) in \cref{fig:comparison}) to further improve the transferability of adversarial examples. However, due to the different characteristics of each image, the most effective combination of image transformations for each image should also be different. %All existing works lack in-depth analysis on better image transformation for improving the attack transferability.

  To solve the problem mentioned above, we propose a novel architecture called Adaptive Image Transformation Learner (AITL), which incorporates different image transformation operations into a unified framework to adaptively select the most effective combination of input transformations towards each image for improving the transferability of adversarial examples. Specifically, AITL consists of encoder and decoder models to convert discrete image transformation operations into continuous feature embeddings, as well as a predictor, which can predict the attack success rate evaluated on black-box models when incorporating the given image transformations into MIFGSM~\cite{dong2018boosting}. After the AITL is well-trained, we optimize the continuous feature embeddings of the image transformation through backpropagation by maximizing the attack success rate, and then use the decoder to obtain the optimized transformation operations. The adaptive combination of image transformations is used to replace the fixed combinational operations in existing methods (as shown in (d) of \cref{fig:comparison}). The subsequent attack process is similar to the mainstream gradient-based attack method~\cite{kurakin2017adversarial,dong2018boosting}. 

  Extensive experiments on ImageNet~\cite{russakovsky2015imagenet} demonstrate that our method not only significantly improves attack success rates on normally trained models, but also shows great effectiveness in attacking various defense models. Especially, we compare our attack method with the combination of state-of-the-art methods~\cite{dong2018boosting,xie2019improving,lin2020nesterov,yang2021adversarial,wang2021admix} against eleven advanced defense methods and achieve a significant improvement of 15.88\% and 5.87\% on average under the single model setting and the ensemble of multiple models setting, respectively. In addition, we conclude that \verb|Scale| is the most effective operation, and geometry-based image transformations (\eg, resizing, rotation, shear) can bring more improvement on the transferability of the adversarial examples, compared to other color-based image transformations (\eg, brightness, sharpness, saturation).

  We summarize our main contributions as follows:

  1. Unlike the fixed combinational transformation used in existing works, we incorporate different image transformations into a unified framework to adaptively select the most effective combination of image transformations for the specific image.

  2. We propose a novel architecture called Adaptive Image Transformation Learner (AITL), which elaborately converts discrete transformations into continuous embeddings and further adopts backpropagation to achieve the optimal solutions, \ie, a combination of effective image transformations for each image.

  3. We conclude that \verb|Scale| is the most effective operation, and geometry-based image transformations are more effective than other color-based image transformations to improve the transferability of adversarial examples.

%-------------------------------------------------------------------------
\section{Related Work}
\label{sec:related}
\subsection{Adversarial Attack}
  The concept of adversarial example is first proposed by Szegedy \etal~\cite{szegedy2014intriguing}. The methods in adversarial attack can be classified as different categories according to the amount of information to the target model the adversary can access, \ie, white-box attack~\cite{goodfellow2015explaining,moosavi2016deepfool,madry2018towards,athalye2018obfuscated,athalye2018synthesizing,croce2020reliable,tramer2020on,duan2021advdrop}, query-based black-box attack~\cite{chen2017zoo,uesato2018adversarial,ilyas2018black-box,li2019nattack,cheng2019improving,du2020query-efficient,ma2021simulating} and transfer-based black-box attack~\cite{dong2018boosting,xie2019improving,dong2019evading,wu2020skip,li2020towards,guo2020backpropagating,lin2020nesterov,wang2021enhancing,wu2021improving}. Since our work focuses on the area of transfer-based black-box attacks, we mainly introduce the methods of transfer-based black-box attack in detail.

  The adversary in transfer-based black-box attack can not access any information about the target model, which only utilizes the transferability of adversarial example~\cite{goodfellow2015explaining} to conduct the attack on the target model. The works in this task can be divided into two main categories, \ie, modified gradient updates and input transformations.
  
  In the branch of modified gradient updates, Dong \etal~\cite{dong2018boosting} first propose MIFGSM to stabilize the update directions with a momentum term to improve the transferability of adversarial examples. Lin \etal~\cite{lin2020nesterov} propose the method of NIM, which adapts Nesterov accelerated gradient into the iterative attacks. Zou \etal~\cite{zou2020making} propose an Adam~\cite{kingma2015adam} iterative fast gradient tanh method (AI-FGTM) to generate indistinguishable adversarial examples with high transferability. Besides, Yang \etal~\cite{yang2021adversarial} absorb the AdaBelief optimizer into the update of the gradient and propose ABI-FGM to further boost the success rates of adversarial examples for black-box attacks. Recently, Wang \etal~propose the techniques of variance tuning~\cite{wang2021enhancing} and enhanced momentum~\cite{wang2021boosting} to further enhance the class of iterative gradient-based attack methods.

  In the branch of various input transformations, Xie \etal~\cite{xie2019improving} propose DIM, which applies random resizing to the inputs at each iteration of I-FGSM~\cite{kurakin2017adversarial} to alleviate the overfitting on white-box models. Dong \etal~\cite{dong2019evading} propose a translation-invariant attack method, called TIM, by optimizing a perturbation over an ensemble of translated images. Lin \etal~\cite{lin2020nesterov} also leverage the scale-invariant property of deep learning models to optimize the adversarial perturbations over the scale copies of the input images.
  Further, Crop-Invariant attack Method (CIM) is proposed by Yang \etal~\cite{yang2021adversarial} to improve the transferability of adversarial. Contemporarily, inspired by mixup~\cite{zhang2018mixup}, Wang \etal~\cite{wang2021admix} propose Admix to calculate the gradient on the input image admixed with a small portion of each add-in image while using the original label of the input, to craft more transferable adversaries. Besides, Wu \etal~\cite{wu2021improving} propose ATTA method, which improves the robustness of synthesized adversarial examples via an adversarial transformation network. Recently, Yuan \etal~\cite{yuan2021automa} propose AutoMA to find the strong model augmentation policy by the framework of reinforcement learning. The works most relevant to ours are AutoMA~\cite{yuan2021automa} and ATTA~\cite{wu2021improving} and we give a brief discussion on the differences between our work and theirs in~\cref{sec:discussion}.

\subsection{Adversarial Defense}
  To boost the robustness of neural networks and defend against adversarial attacks, numerous methods of adversarial defense have been proposed.
  
  Adversarial training~\cite{goodfellow2015explaining,kurakin2017adversarial,madry2018towards} adds the adversarial examples generated by several methods of adversarial attack into the training set, to boost the robustness of models. Although effective, the problems of huge computational cost and overfitting to the specific attack pattern in adversarial training receive increasing concerns. Several follow-up works~\cite{rozsa2016adversarial,madry2018towards,tramer2018ensemble,wang2019on,dong2020adversarial,pang2020boosting,wu2020adversarial,wong2020fast} aim to solve these problems.
  Another major approach is the method of input transformation, which preprocesses the input to mitigate the adversarial effect ahead, including JPEG compression~\cite{guo2018countering,liu2019feature}, denoising~\cite{liao2018defense}, random resizing~\cite{xie2018mitigating}, bit depth reduction~\cite{xu2018feature} and so on. 
  Certified defense~\cite{katz2017reluplex,xiao2019training,croce2020provable,jia2020certified} attempts to provide a guarantee that the target model can not be fooled within a small perturbation neighborhood of the clean image.
  Moreover, Jia \etal~\cite{jia2019comdefend} utilize an image compression model to defend the adversarial examples. Naseer \etal~\cite{naseer2020a} propose a self-supervised adversarial training mechanism in the input space to combine the benefit of both the adversarial training and input transformation method. The various defense methods mentioned above help to improve the robustness of the model.

%-------------------------------------------------------------------------
\section{Method}
\label{sec:method}
  In this section, we first give the definition of the notations in the task. And then we introduce our proposed Adaptive Image Transformation Learner (AITL), which can adaptively select the most effective combination of image transformations used during the attack to improve the transferability of generated adversarial examples.
\subsection{Notations}
  Let $x\in \mathcal{X}$ denote a clean image from a dataset of $\mathcal{X}$, and $y \in \mathcal{Y}$ is the corresponding ground truth label. Given a source model $f$ with parameters $\theta$, the objective of adversarial attack is to find the adversarial example $x^{adv}$ that satisfies: 
  \begin{equation}
    f(x^{adv}) \neq y, \quad s.t. \|x-x^{adv}\|_\infty \leq \epsilon,
  \end{equation}
  where $\epsilon$ is a preset parameter to constrain the intensity of the perturbation. In implementation, most gradient-based adversaries utilize the method of maximizing the loss function to iteratively generate adversarial examples. We here take the widely used method of MIFGSM~\cite{dong2018boosting} as an example:
  \begin{gather}
    g_{t+1} = \mu \cdot g_t + \frac{\nabla_{x_t^{adv}} J(f(x_t^{adv}), y)}{\|\nabla_{x_t^{adv}} J(f(x_t^{adv}), y)\|_1}, \label{equ:g} \\
    x_{t+1}^{adv} = x_t^{adv} + \alpha \cdot sign (g_{t+1}), \\
    g_0 = 0, \quad x_0^{adv} = x,
  \end{gather}
  where $g_t$ is the accumulated gradients, $x_t^{adv}$ is the generated adversarial example at the time step $t$, $J(\cdot)$ is the loss function used in classification models (\ie, the cross entropy loss), $\mu$ and $\alpha$ are hyperparameters.

\subsection{Overview of AITL}
  Existing works of input-transformation-based methods have studied the influence of some input transformations on the transferability of adversarial examples. These methods can be combined with the MIFGSM~\cite{dong2018boosting} method and can be summarized as the following paradigm, where the \cref{equ:g} is replaced by:
  \begin{equation}
    \label{equ:T}
    g_{t+1} = \mu \cdot g_t + \frac{\nabla_{x_t^{adv}} J({f(\color{red}T(}x_t^{adv}{\color{red})}), y)}{\|\nabla_{x_t^{adv}} J(f({\color{red}T(}x_t^{adv}{\color{red})}), y)\|_1},
  \end{equation}
  where $T$ represents different input transformation operations in different method (\eg, resizing in DIM~\cite{xie2019improving}, translation in TIM~\cite{dong2019evading}, scaling in SIM~\cite{lin2020nesterov}, cropping in CIM~\cite{yang2021adversarial}, mixup in Admix~\cite{wang2021admix}).
  
  Although existing methods improve the transferability of adversarial examples to a certain extent, almost all of these methods only utilize different image transformations respectively and haven't systematically studied which transformation operation is more suitable. Also, these methods haven't considered the characteristic of each image, but uniformly adopt a fixed transformation method for all images, which is not reasonable in nature and cannot maximize the transferability of the generated adversarial examples.
  
  In this paper, we incorporate different image transformation operations into a unified framework and utilize an Adaptive Image Transformation Learner (AITL) to adaptively select the suitable input transformations towards different input images (as shown in (d) of \cref{fig:comparison}). This unified framework can analyze the impact of different transformations on the generated adversarial examples.

  Overall, our method consists of two phases, \ie, the phase of training AITL to learn the relationship between various image transformations and the corresponding attack success rates, and the phase of generating adversarial examples with well-trained AITL. During the training phase, we conduct encoder and decoder networks, which can convert the discretized image transformation operations into continuous feature embeddings. In addition, a predictor is proposed to predict the attack success rate in the case of the original image being firstly transformed by the given image transformation operations and then attacked with the method of MIFGSM~\cite{dong2018boosting}. After the training of AITL is finished, we maximize the attack success rate to optimize the continuous feature embeddings of the image transformation through backpropagation, and then use the decoder to obtain the optimal transformation operations specific to the input, and incorporate it into MIFGSM to conduct the actual attack.
  
  In the following two subsections, we will introduce the two phases mentioned above in detail, respectively.

\begin{figure*}[t]
  \centering
  \includegraphics[width=0.9\linewidth]{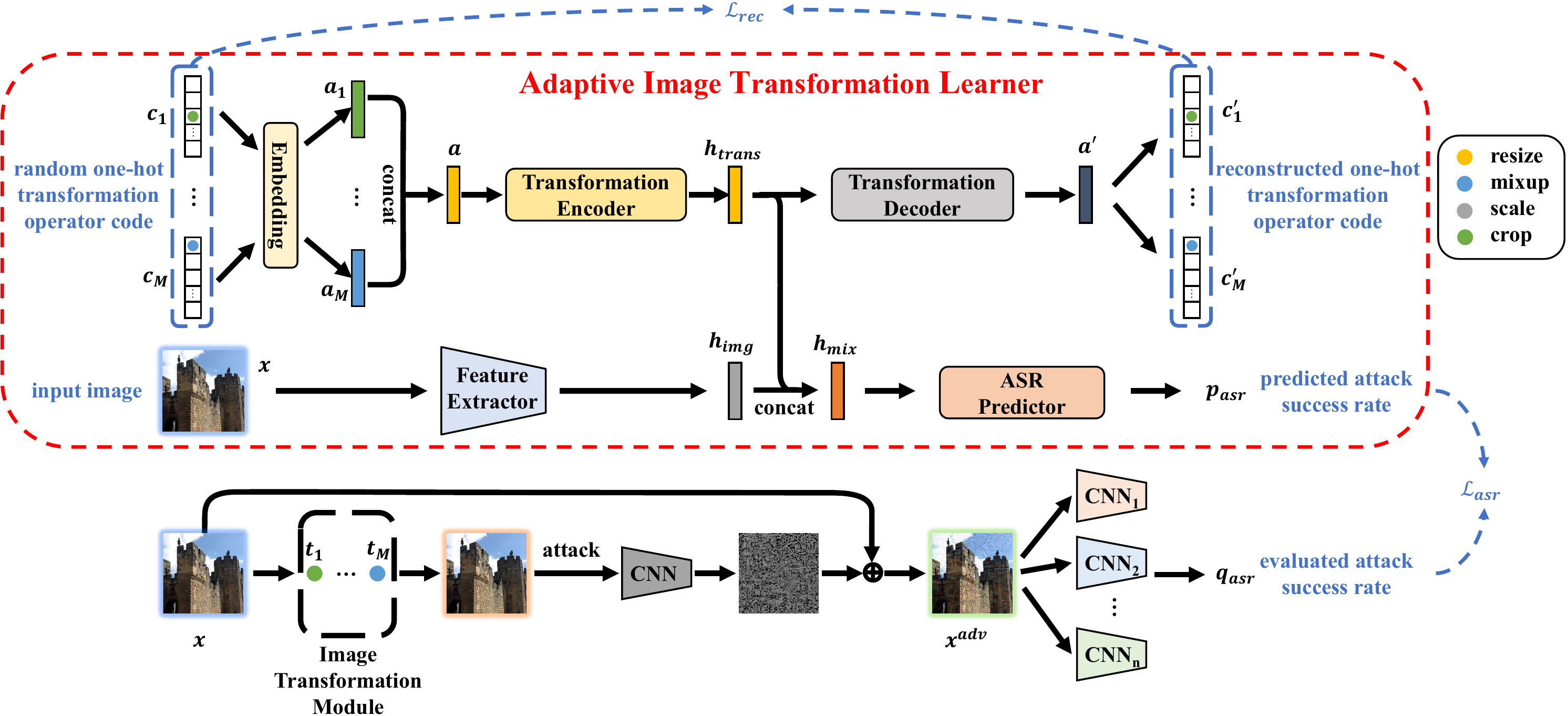}
  % \vspace{-7mm}
  \caption{The diagram of Adaptive Image Transformation Learner in the process of training}
  \label{fig:training}
  % \vspace{-5mm}
\end{figure*}

\subsection{Training AITL}
  The overall process of training AITL is shown in \cref{fig:training}.
  We first randomly select $M$ image transformations $t_1, t_2, \cdots, t_M$ from the image transformation operation zoo (including both geometry-based and color-based operations, for details please refer to \cref{sec:setup:operation}) based on uniform distribution to compose an image transformation combination.
  We then discretize different image transformations by encoding them into one-hot vectors $c_1, c_2, \cdots, c_M$ (\eg, $[1,0,0,\cdots]$ represents resizing, $[0, 1, 0, \cdots]$ represents scaling).
  An embedding layer then converts different transformation operations into their respective feature vectors, which are concatenated into an integrated input transformation feature vector $a$:
  \begin{gather}
    a_1, a_2, \cdots, a_M = Embedding(c_1, c_2, \cdots, c_M), \\
    a = Concat(a_1, a_2, \cdots, a_M).
  \end{gather}
  The integrated input transformation feature vector then goes through a transformation encoder $f_{en}$ and decoder $f_{de}$ in turn, so as to learn the continuous feature embeddings $h_{trans}$ in the intermediate layer: 
  \begin{gather}
    h_{trans} = f_{en}(a), \\
    a' = f_{de}(h_{trans}).
  \end{gather}
  The resultant decoded feature $a'$ is then utilized to reconstruct the input transformation one-hot vectors:
  \begin{equation}
    c_1', c_2' ,\cdots, c_M' = FC(a'),
  \end{equation}
  where $FC$ represents a fully connected layer with multiple heads, each represents the reconstruction of an input image transformation operation.
  On the other hand, a feature extractor $f_{img}$ is utilized to extract the image feature of the original image $h_{img}$, which is concatenated with the continuous feature embeddings of image transformation combination $h_{trans}$:
  \begin{gather}
    h_{img} = f_{img}(x), \\
    h_{mix} = Concat(h_{trans}, h_{img}).
  \end{gather}
  Then the mixed feature is used to predict the attack success rate $p_{asr}$ through an attack success rate predictor $f_{pre}$ in the case of the original image being firstly transformed by the input image transformation combination and then attacked with the method of MIFGSM:
  \begin{equation}
    p_{asr} = f_{pre}(h_{mix}).
  \end{equation}
 
  \textbf{Loss Functions.} The loss function used to train the network consists of two parts. The one is the reconstruction loss $\mathcal{L}_{rec}$ to constrain the reconstructed image transformation operations $c_1', c_2', \cdots, c_M'$ being consistent with the input image transformation operations $c_1, c_2, \cdots, c_M$:
  \begin{equation}
    \mathcal{L}_{rec} = -\sum_{i=1}^{M} c_i^T \log c_{i}',
  \end{equation}
  where $T$ represents the transpose of a vector.
  The other one is the prediction loss $\mathcal{L}_{asr}$, which aims to ensure that the attack success rate predicted by the ASR predictor $p_{asr}$ is close to the actual attack success rate $q_{asr}$.
  \begin{equation}
    \mathcal{L}_{asr} = \| p_{asr} - q_{asr} \|_2.
  \end{equation}
  The actual attack success rate $q_{asr}$ is achieved by evaluating the adversarial example $x^{adv}$, which is generated through replacing the fixed transformation operations in existing methods by the given input image transformation combination (\ie, $T=t_M \circ \cdots \circ t_2 \circ t_1$ in \cref{equ:T}) on $n$ black-box models $f_1, f_2, \cdots, f_n$ (as shown in the bottom half in \cref{fig:training}):
  \begin{equation}
    q_{asr} = \frac{1}{n} \sum_{i=1}^n \mathbbm{1}(f_i(x^{adv}) \neq y).
  \end{equation}
  And the total loss function is the sum of above introduced two items:
  \begin{equation}
    \mathcal{L}_{total} = \mathcal{L}_{rec} + \mathcal{L}_{asr}.
  \end{equation}
  The entire training process is summarized in \cref{alg:training} in the appendix.

\begin{figure*}[t]
  \centering
  \includegraphics[width=0.9\linewidth]{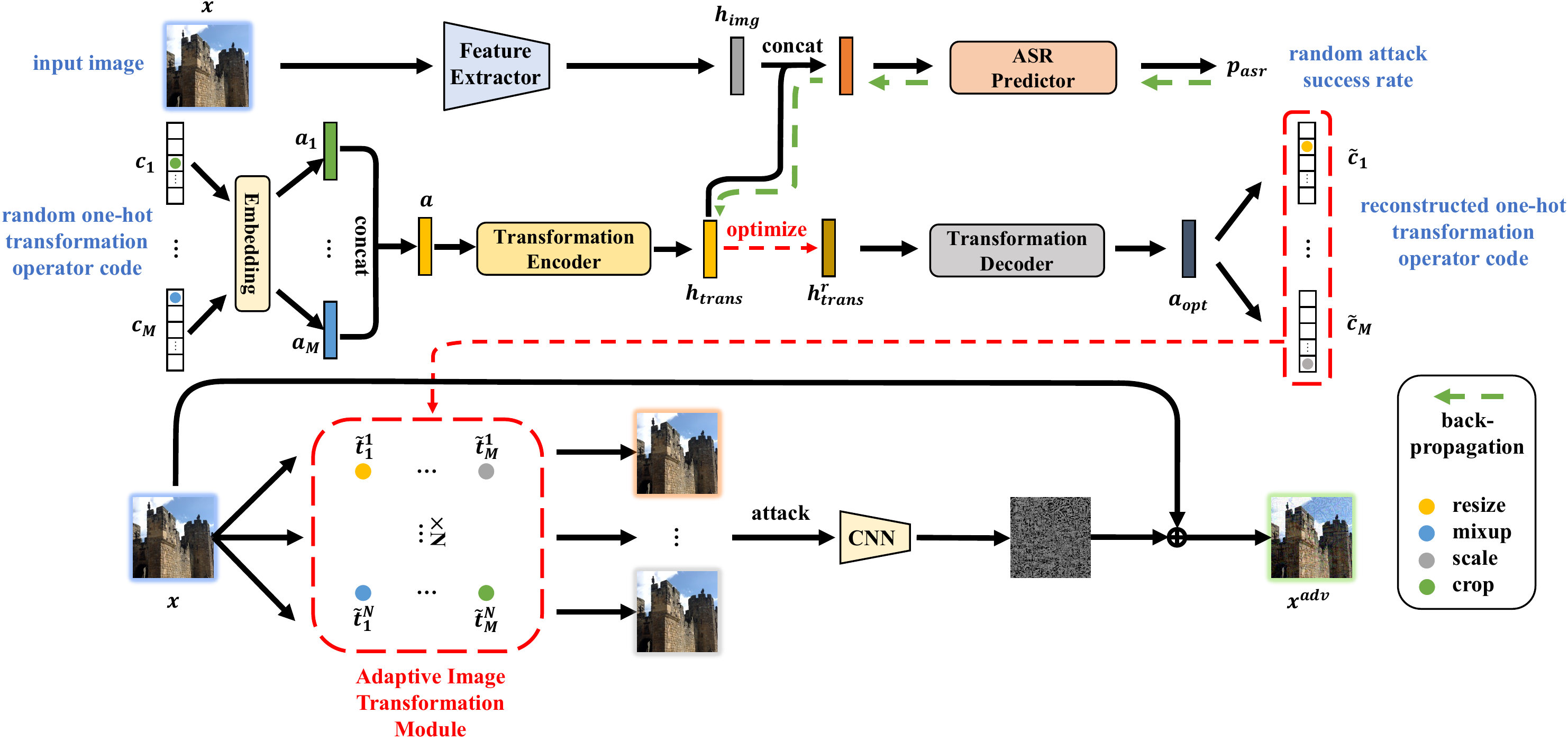}
  % \vspace{-7mm}
  \caption{The process of generating adversarial examples with Adaptive Image Transformation Learner}
  \label{fig:inference}
  % \vspace{-8mm}
\end{figure*}

\subsection{Generating Adversarial Examples with AITL}
  When the training of Adaptive Image Transformation Learner is finished, it can be used to adaptively select the appropriate combination of image transformations when conducting adversarial attacks against any unknown model. The process has been shown in \cref{fig:inference}.

  For an arbitrary input image, we can not identify the most effective input transformation operations that can improve the transferability of generated adversarial examples ahead. Therefore, we first still randomly sample $M$ initial input transformation operations $t_1, t_2, \cdots, t_M$, and go through a forward pass in AITL to get the predicted attack success rate $p_{asr}$ corresponding to the input transformation operations. Then we iteratively optimize the image transformation feature embedding $h_{trans}$ by maximizing the predicted attack success rate for $r$ times:
  \begin{gather}
    h_{trans}^{t+1} = h_{trans}^{t} + \gamma \cdot \nabla_{h_{trans}^{t}} p_{asr}, \\
    h_{trans}^{0} = h_{trans},
  \end{gather}
  where $\gamma$ is the step size in each optimizing step. Finally we achieve the optimized image transformation feature embedding $h_{trans}^r$. Then we utilize the pre-trained decoder to convert the continuous feature embedding into specific image transformation operations:
  \begin{gather}
    a_{opt} = f_{de}(h_{trans}^r), \\
    \tilde{c}_1, \tilde{c}_2, \cdots, \tilde{c}_M = FC(a_{opt}).
  \end{gather}
  The resultant image transformation operations $\tilde{c}_1, \tilde{c}_2, \cdots, \tilde{c}_M$ achieved by AITL are considered to be the most effective combination of image transformations for improving the transferability of generated adversarial example towards the specific input image. Thus we utilize these image transformation operations to generate adversarial examples. When combined with MIFGSM~\cite{dong2018boosting}, the whole process can be summarized as:
  \begin{gather}
    \tilde{c}_1, \tilde{c}_1, \cdots, \tilde{c}_M = AITL(x), \\
    g_{t+1} = \mu \cdot g_t + \frac{\nabla_{x_t^{adv}} J(f(\tilde{c}_M \circ \cdots \circ \tilde{c}_1(x_t^{adv})), y)}{\|\nabla_{x_t^{adv}} J(f(\tilde{c}_M \circ \cdots \circ \tilde{c}_1(x_t^{adv})), y)\|_1}, \\
    x_{t+1}^{adv} = x_t^{adv} + \alpha \cdot sign (g_{t+1}), \\
    g_0 = 0, \quad x_0^{adv} = x.
  \end{gather}

  Since the random image transformation operations contain randomness (\eg, the degree in rotation, the width and height in resizing), existing works~\cite{lin2020nesterov,yang2021adversarial,wang2021admix} conduct these transformation operations multiple times in parallel during each step of the attack to alleviate the impact of the instability caused by randomness on the generated adversarial examples (as shown in (b) of \cref{fig:comparison}). Similar to previous works, we also randomly sample the initial image transformation combination multiple times, and then optimize them to obtain the optimal combination of image transformation operations respectively. The several optimal image transformation combinations are used in parallel to generate adversarial examples (as shown in the bottom half in \cref{fig:inference}). The entire process of using AITL to generate adversarial examples is formally summarized in \cref{alg:inference} in the appendix.

  The specific network structure of the entire framework is shown in \cref{sec:setup:network}. The ASR predictor, Transformation Encoder and Decoder in our AITL consist of only a few FC layers. During the iterative attack, our method only needs to infer once before the first iteration. So our AITL is a lightweight method, and the extra cost compared to existing methods is negligible.

%------------------------------------------------------------------------
\section{Experiments}
\label{sec:exper}
  In this section, we first introduce the settings in the experiments in \cref{sec:exper:setting}. Then we demonstrate the results of our proposed AITL method on the single model attack and an ensemble of multiple models attack, respectively in \cref{sec:exper:attack}. We also analyze the effectiveness of different image transformation methods in \cref{sec:exper:analysis}. In \cref{sec:extra}, more extra experiments are provided, including the attack success rate under different perturbation budgets, the influence of some hyperparameters, more experiments on the single model attack, the results of AITL combined with other base attack methods and the visualization of generated adversarial examples.

\subsection{Settings}
\label{sec:exper:setting}
  \textbf{Dataset.} We use two sets of subsets\footnote{\url{https://github.com/cleverhans-lab/cleverhans/tree/master/cleverhans_v3.1.0/examples/nips17_adversarial_competition/dataset}}\textsuperscript{,}\footnote{\url{https://drive.google.com/drive/folders/1CfobY6i8BfqfWPHL31FKFDipNjqWwAhS}} in the ImageNet dataset~\cite{russakovsky2015imagenet} to conduct experiments. Each set contains 1000 images, covering almost all categories in ImageNet, which has been widely used in previous works~\cite{dong2018boosting,dong2019evading,lin2020nesterov}. All images have the size of $299\times 299 \times 3$. In order to make a fair comparison with other methods, we use the former subset to train the AITL model, and evaluate all methods on the latter one. 

  \textbf{Models.} In order to avoid overfitting of the AITL model and ensure the fairness of the experimental comparison, we use completely different models to conduct experiments during the training and evaluation of the AITL model. During the training, we totally 11 models to provide the attack success rate corresponding to the input transformation, including 10 normally trained and 1 adversarially trained models. During the evaluation, we use 7 normally trained models, 3 adversarially trained models, and another 8 stronger defense models to conduct the experiments. The details are provided in \cref{sec:setup:model}. 

  \textbf{Baselines.} Several input-transformation-based black-box attack methods (\eg, DIM~\cite{xie2019improving}, TIM~\cite{dong2019evading}, SIM~\cite{lin2020nesterov}, CIM~\cite{yang2021adversarial}, Admix~\cite{wang2021admix}, AutoMA~\cite{yuan2021automa}) are utilized to compare with our proposed method. Unless mentioned specifically, we combine these methods with MIFGSM~\cite{dong2018boosting} to conduct the attack. In addition, we also combine these input-transformation-based methods together to form the strongest baseline, called Admix-DI-SI-CI-MIFGSM (as shown in (c) of \cref{fig:comparison}, ADSCM for short). Moreover, we also use a random selection method instead of the AITL to choose the combination of image transformations used in the attack, which is denoted as \verb|Random|. The details of these baselines are provided in \cref{sec:setup:baseline}.

  \textbf{Image Transformation Operations.} Partially referencing from~\cite{cubuk2019autoaugment,cubuk2020randaugment}, we totally select 20 image transformation operations as candidates, including \verb|Admix|, \verb|Scale|, \verb|Admix-and-Scale|, \verb|Brightness|, \verb|Color|, \verb|Contrast|, \verb|Sharpness|, \verb|Invert|, \verb|Hue|, \verb|Saturation|, \verb|Gamma|, \verb|Crop|, \verb|Resize|, \verb|Rotate|, \verb|ShearX|, \verb|ShearY|, \verb|TranslateX|, \verb|TranslateY|, \verb|Reshape|, \verb|Cutout|. The details of these operations are provided in \cref{sec:setup:operation}, including the accurate definitions and specific parameters in the random transformations.

  \textbf{Implementation Details.} We train the AITL model for 10 epochs. The batch size is 64, and the learning rate $\beta$ is set to 0.00005. The detailed network structure of AITL is introduced in \cref{sec:setup:network}. The maximum adversarial perturbation $\epsilon$ is set to 16, with an iteration step $T$ of 10 and step size $\alpha$ of 1.6. The number of iterations during optimizing image transformation features $r$ is set to 1 and the corresponding step size $\gamma$ is 15. The number of image transformation operations used in a combination $M$ is set to 4 (the same number as the transformations used in the strongest baseline ADSCM for a fair comparison). Also, for a fair comparison of different methods, we control the number of repetitions per iteration in all methods to 5 ($m$ in SIM~\cite{lin2020nesterov}, $m_2$ in Admix~\cite{wang2021admix}, $m$ in AutoMA~\cite{yuan2021automa} and $N$ in our AITL).

\begin{table*}[!t]
  \centering
  \caption{Attack success rates (\%) of adversarial attacks against 7 normally trained models and 11 defense models under \textbf{single model} setting. The adversarial examples are crafted on Incv3. $^*$ indicates the white-box model. $^\dagger$ The results of AutoMA~\cite{yuan2021automa} are cited from their original paper}
    \text{(a) The evaluation against 7 normally trained models}
    \resizebox{0.75\textwidth}{!}{
      \begin{tabular}{c|cccccccc}
        \hline
        \noalign{\smallskip}
        & Incv3$^*$ & Incv4 & IncResv2 & Resv2-101 & Resv2-152 & PNASNet & NASNet \\
        \noalign{\smallskip}
        \hline
        \noalign{\smallskip}
        MIFGSM~\cite{dong2018boosting} & \textbf{100}& 52.2& 50.6& 37.4& 35.6& 42.2& 42.2\\
        DIM~\cite{xie2019improving} & 99.7& 78.3& 76.3& 59.6& 59.9& 64.6& 66.2 \\
        SIM~\cite{lin2020nesterov} & \textbf{100}& 84.5& 81.3& 68.0& 65.3& 70.8& 73.6 \\
        CIM~\cite{yang2021adversarial} & \textbf{100}& 85.1& 81.6& 58.1& 57.4& 65.7& 66.7 \\
        Admix~\cite{wang2021admix} & 99.8& 69.5& 66.5& 55.3& 55.4& 60.0& 62.7 \\
        ADSCM & \textbf{100}& 87.9& 86.1& 75.8& 76.0& 80.9& 82.2 \\
        Random & \textbf{100}& 94.0& 92.0& 79.7& 80.0& 84.6& 85.5 \\
        AutoMA$^\dagger$~\cite{yuan2021automa} & 98.2& 91.2& 91.0& 82.5& - & - & - \\
        AITL (ours) & 99.8& \textbf{95.8}& \textbf{94.1}& \textbf{88.8}& \textbf{90.1} & \textbf{94.1} & \textbf{94.0} \\
        \noalign{\smallskip}
        \hline
        \noalign{\smallskip}
        AutoMA-TIM$^\dagger$~\cite{yuan2021automa} & 97.5& 80.7& 74.3& 69.3& - & -& -  \\
        AITL-TIM (ours) & \textbf{99.8}& \textbf{93.4}& \textbf{92.1}& \textbf{91.9}& \textbf{92.2} & \textbf{93.8}& \textbf{94.6}  \\
        \noalign{\smallskip}
        \hline
      \end{tabular}
    }
    \text{(b) The evaluation against 11 defense models}
    \resizebox{\textwidth}{!}{
      \begin{tabular}{c|ccccccccccc}
        \hline
        \noalign{\smallskip}
        & Incv3$_{\texttt{ens3}}$ & Incv3$_{\texttt{ens4}}$ & IncResv2$_{\texttt{ens}}$ & HGD & R\&P & NIPS-r3 & Bit-Red & JPEG & FD & ComDefend & RS \\
        \noalign{\smallskip}
        \hline
        \noalign{\smallskip}
        MIFGSM~\cite{dong2018boosting} & 15.6& 15.2& 6.4& 5.8& 5.6& 9.3& 18.5& 33.3& 39.0& 28.1& 16.8 \\
        DIM~\cite{xie2019improving} & 31.0& 29.2& 13.4& 15.8& 14.8& 24.6& 26.8& 59.3& 45.8& 48.3& 21.8 \\
        SIM~\cite{lin2020nesterov} & 37.5& 35.0& 18.8& 16.8& 18.3& 26.8& 31.0& 66.9& 52.1& 55.9& 24.1 \\
        CIM~\cite{yang2021adversarial} & 33.3& 30.0& 15.9& 20.4& 16.4& 25.7& 26.8& 62.2& 46.3& 44.9& 21.2 \\
        Admix~\cite{wang2021admix} & 27.5& 27.0& 14.3& 11.6& 12.6& 19.8& 28.4& 51.2& 48.8& 44.0& 22.0\\
        ADSCM & 49.3& 46.9& 27.0& 33.1& 28.5& 40.5& 39.0& 73.0& 60.4& 65.5& 32.8\\
        Random & 49.8& 46.7& 24.5& 29.2& 26.4& 42.2& 36.3& 81.4& 57.4& 69.6& 29.6 \\
        AutoMA$^\dagger$~\cite{yuan2021automa} &49.2 & 49.0& 29.1& -&-&-&-&-&-&-&- \\
        AITL (ours) & \textbf{69.9}& \textbf{65.8}& \textbf{43.4}& \textbf{50.4}& \textbf{46.9}& \textbf{59.9}& \textbf{51.6}& \textbf{87.1}& \textbf{73.0}& \textbf{83.2}& \textbf{39.5} \\
        \noalign{\smallskip}
        \hline
        \noalign{\smallskip}
        AutoMA-TIM$^\dagger$~\cite{yuan2021automa} &74.8 & 74.3& 63.6& 65.7&62.9&68.1&-&-&\textbf{84.7}&-&- \\
        AITL-TIM (ours) & \textbf{81.3}& \textbf{78.9}& \textbf{69.1}& \textbf{75.1}& \textbf{64.7}&\textbf{74.6}&\textbf{60.9}&\textbf{87.8}&83.8&\textbf{85.6}&\textbf{55.4} \\
        \noalign{\smallskip}
        \hline
      \end{tabular}
    }
  \label{tab:single}
\end{table*}

\subsection{Compared with the State-of-the-art Methods}
\label{sec:exper:attack}

\subsubsection{Attack on the Single Model.}
  We use Inceptionv3~\cite{szegedy2016rethinking} model as the white-box model to conduct the adversarial attack, and evaluate the generated adversarial examples on both normally trained models and defense models. As shown in \cref{tab:single}, comparing various existing input-transformation-based methods, our proposed AITL significantly improves the attack success rates against various black-box models. Especially for the defense models, although it is relatively difficult to attack successfully, our method still achieves a significant improvement of 15.88\% on average, compared to the strong baseline (Admix-DI-SI-CI-MIFGSM). It demonstrates that, compared with the fixed image transformation combination, adaptively selecting combinational image transformations for each image can indeed improve the transferability of adversarial examples. Also, when compared to AutoMA~\cite{yuan2021automa}, our AITL achieves a distinct improvement, which shows that our AITL model achieves better mapping between discrete image transformations and continuous feature embeddings. More results of attacking other models are available in \cref{sec:add:single}. Noting that the models used for evaluation here are totally different from the models used when training the AITL, our method shows great cross model transferability to conduct the successful adversarial attack.

\subsubsection{Attack on the Ensemble of Multiple Models.}
  We use the ensemble of four models, \ie, Inceptionv3~\cite{szegedy2016rethinking}, Inceptionv4~\cite{szegedy2017inception}, Inception-ResNetv2~\cite{szegedy2017inception} and ResNetv2-101~\cite{he2016identity}, as the white-box models to conduct the adversarial attack. As shown in \cref{tab:multiple}, compared with the fixed image transformation method, our AITL significantly improves the attack success rates on various models. Although the strong baseline ADSCM has achieved relatively high attack success rates, our AITL still obtains an improvement of 1.44\% and 5.87\% on average against black-box normally trained models and defense models, respectively. Compared to AutoMA~\cite{yuan2021automa}, our AITL also achieves higher attack success rates on defense models, which shows the superiority of our proposed novel architecture.

\begin{table*}[!t]
  \centering
  \caption{Attack success rates (\%) of adversarial attacks against 7 normally trained models and 11 defense models under \textbf{multiple models} setting. The adversarial examples are crafted on the ensemble of Incv3, Incv4, IncResv2 and Resv2-101. $^*$ indicates the white-box model. $^\dagger$ The results of AutoMA~\cite{yuan2021automa} are cited from their original paper}
    \centering
    \text{(a) The evaluation against 7 normally trained models}
    \resizebox{0.75\textwidth}{!}{
      \begin{tabular}{c|ccccccc}
        \hline
        \noalign{\smallskip}
        & Incv3$^*$ & Incv4$^*$ & IncResv2$^*$ & Resv2-101$^*$ & Resv2-152 & PNASNet & NASNet \\
        \noalign{\smallskip}
        \hline
        \noalign{\smallskip}
        MIFGSM~\cite{dong2018boosting} & \textbf{100}& 99.6& 99.7& \textbf{98.5}& 86.8& 79.4& 81.2\\
        DIM~\cite{xie2019improving} & 99.5& 99.4& 98.9& 96.9& 92.0& 91.3& 92.1\\
        SIM~\cite{lin2020nesterov} & 99.9& 99.1& 98.3& 93.2& 91.7& 90.9& 91.9\\
        CIM~\cite{yang2021adversarial} & 99.8& 99.3& 97.8& 90.6& 88.5& 88.2& 90.9\\
        Admix~\cite{wang2021admix} & 99.9&99.5 &98.2 &95.4 &89.3 &88.1 &90.0 \\
        ADSCM &99.8 &99.3 &99.2 &96.9 &96.0 &88.1 &90.0 \\
        Random & \textbf{100}& 99.4& 98.9& 96.9& 94.3& 94.4& 95.0\\
        AITL (ours) & 99.9 &\textbf{99.7}&\textbf{99.9}&97.3& \textbf{96.6}&\textbf{97.7}& \textbf{97.8} \\
        \noalign{\smallskip}
        \hline
      \end{tabular}
    }
    \text{(b) The evaluation against 11 defense models}
    \resizebox{\textwidth}{!}{
      \begin{tabular}{c|ccccccccccc}
        \hline
        \noalign{\smallskip}
        & Incv3$_{\texttt{ens3}}$ & Incv3$_{\texttt{ens4}}$ & IncResv2$_{\texttt{ens}}$ & HGD & R\&P & NIPS-r3 & Bit-Red & JPEG & FD & ComDefend & RS \\
        \noalign{\smallskip}
        \hline
        \noalign{\smallskip}
        MIFGSM~\cite{dong2018boosting} & 52.4& 47.5& 30.1& 39.2& 31.7& 43.6& 33.8& 76.4& 54.5& 66.8& 29.7\\
        DIM~\cite{xie2019improving} &77.4& 73.1& 54.4& 68.4& 61.2& 73.5& 53.3& 89.8& 71.5& 84.3& 43.1\\
        SIM~\cite{lin2020nesterov} &78.8& 74.4& 59.8& 66.9& 59.0& 70.7& 58.1& 89.0& 73.2& 83.0& 46.6\\
        CIM~\cite{yang2021adversarial} &75.1& 69.7& 54.3& 68.5& 59.1& 70.7& 51.1& 90.2& 68.9& 78.5& 41.1\\
        Admix~\cite{wang2021admix} &67.7& 61.9& 44.8& 51.0&44.8 &57.9 &51.4 &84.6 &69.2 &78.5&42.2 \\
        ADSCM &85.8&82.9 &69.2 &78.7 &74.1 &81.1 &68.1 &94.9 &82.3& 90.8&57.8 \\
        Random & 83.7& 80.2&64.8&73.7&67.3&77.9&65.7&93.0&79.9&88.6&52.3\\
        AITL (ours) & \textbf{89.3}&\textbf{89.0}&\textbf{79.0}&\textbf{85.5}&\textbf{82.3}&\textbf{86.3}&\textbf{74.9}&\textbf{96.2}&\textbf{88.4}&\textbf{93.7}&\textbf{65.7} \\
        \noalign{\smallskip}
        \hline
        \noalign{\smallskip}
        AutoMA-TIM$^\dagger$~\cite{yuan2021automa} &93.0&93.2&90.7&91.2&90.4&92.0&-&-&94.1&-&-\\
        AITL-TIM (ours) &\textbf{93.8}& \textbf{95.3}& \textbf{92.0}&\textbf{93.1}&\textbf{93.7}&\textbf{94.8}&\textbf{80.9}&\textbf{95.0}&\textbf{96.2}&\textbf{95.0}&\textbf{76.9}\\
        \noalign{\smallskip}
        \hline
      \end{tabular}
    }
  \label{tab:multiple}
\end{table*}

\subsection{Analysis on Image Transformation Operations}
\label{sec:exper:analysis}
\begin{figure}[t]
  \centering
  \includegraphics[width=0.5\linewidth]{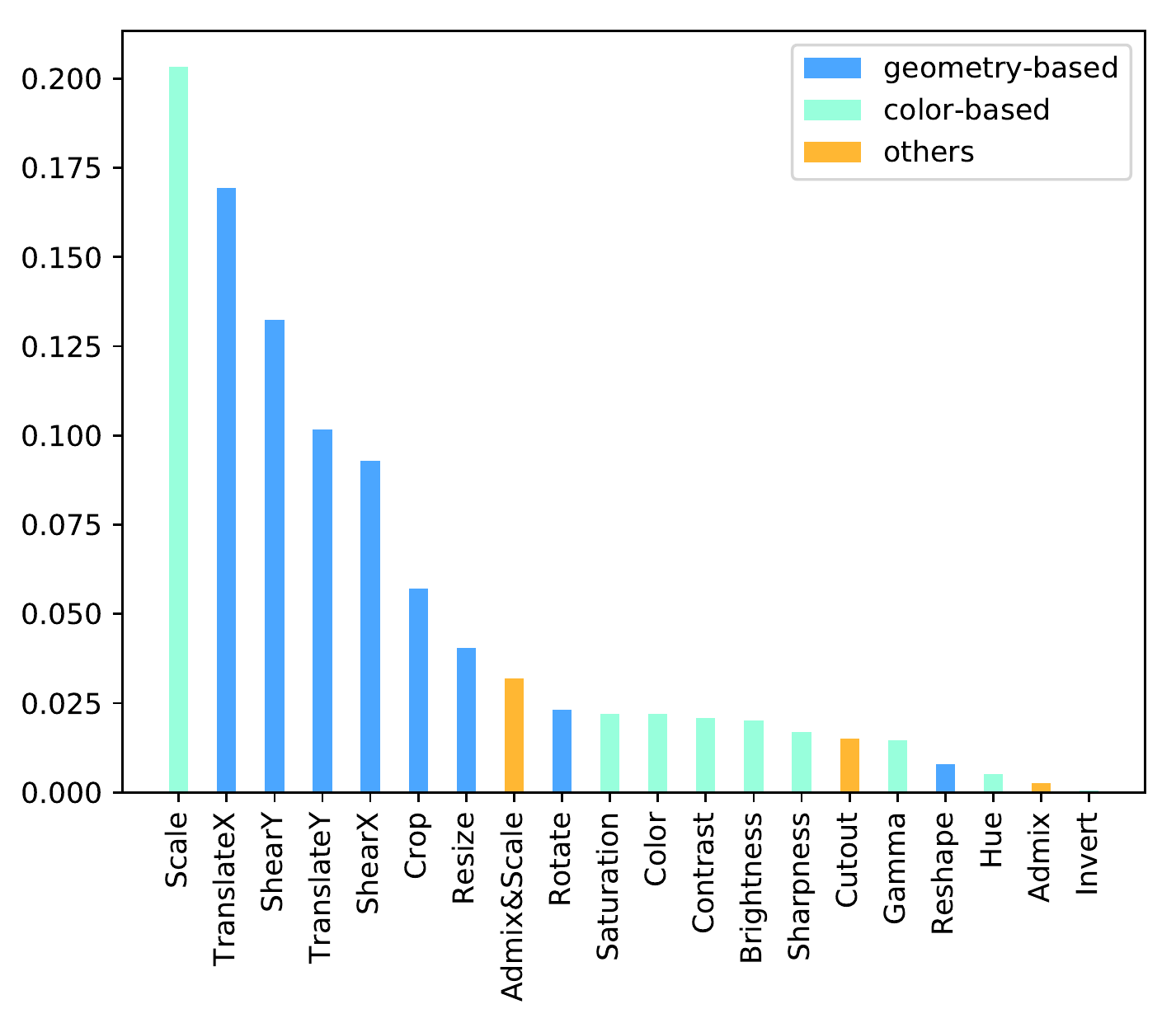}
  \caption{The frequency of various image transformations used in AITL when generating adversarial examples of the 1000 images in ImageNet}
  \label{fig:bar}
\end{figure}

  In order to further explore the effects of different image transformation operations on improving the transferability of adversarial examples, we calculate the frequency of various image transformations used in AITL when generating adversarial examples of the 1000 images in ImageNet. From \cref{fig:bar}, we can clearly see that \verb|Scale| operation is the most effective method within all 20 candidates. Also, we conclude that the geometry-based image transformations are more effective than other color-based image transformations to improve the transferability of adversarial examples.
%-------------------------------------------------------------------------
\section{Conclusion}
\label{sec:conclusion}
  In our work, unlike the fixed image transformation operations used in almost all existing works of transfer-based black-box attack, we propose a novel architecture, called Adaptive Image Transformation Learner (AITL), which incorporates different image transformation operations into a unified framework to further improve the transferability of adversarial examples. By taking the characteristic of each image into consideration, our designed AITL adaptively selects the most effective combination of image transformations for the specific image. Extensive experiments on ImageNet demonstrate that our method significantly improves the attack success rates both on normally trained models and defense models under different settings.

  \textbf{Acknowledgments.} This work is partially supported by National Key R\&D Program of China (No. 2017YFA0700800), National Natural Science Foundation of China (Nos. 62176251 and Nos. 61976219).

\clearpage
% ---- Bibliography ----
%
% BibTeX users should specify bibliography style 'splncs04'.
% References will then be sorted and formatted in the correct style.
%
\bibliographystyle{splncs04}
\bibliography{egbib}

\clearpage
\appendix
\section*{Appendix}

\section{Details of the Settings in the Experiment}
\subsection{Models}
\label{sec:setup:model}
  In order to avoid overfitting of the AITL model and ensure the fairness of the experimental comparison, we use completely different models to conduct experiments during the training and evaluation of the AITL model.
  
  During the process of training AITL, we utilize totally eleven models to provide the attack success rate corresponding to the input transformation, including ten normally trained models (\ie, ResNet-50~\cite{he2016deep}, Xception~\cite{chollet2017xception}, DenseNet-201~\cite{huang2017densely}, VGG-19~\cite{simonyan2014very}, MobileNetv2-1.0~\cite{sandler2018mobilenetv2}, MobileNetv2-1.4~\cite{sandler2018mobilenetv2}, ResNeXt-101~\cite{xie2017aggregated}, SENet-101~\cite{hu2018squeeze}, EfficientNetB4~\cite{tan2019efficientnet} and EfficientNetv2S~\cite{tan2021efficientnetv2}) and one adversarially trained model (\ie, AdvInceptionv3~\cite{tramer2017ensemble}). All models are publicly available\footnote{\url{https://keras.io/api/applications}}\textsuperscript{,}\footlabel{note1}{\url{https://github.com/wowowoxuan/adv_imagenet_models}}. In the process of training Adaptive Image Transformation Learner, we utilize the MobileNetv2-1.0 as the source white-box model (as the grey model in \cref{fig:training}) to generate the adversarial examples, and regard the other models as the target black-box models (as the other colorful models in \cref{fig:training}) to evaluate the corresponding attack success rates, \ie, $q_{asr}$.
  
  During the process of evaluation, we use seven normally trained models (\ie, Inceptionv3 (Incv3)~\cite{szegedy2016rethinking}, Inceptionv4 (Incv4)~\cite{szegedy2017inception}, Inception-ResNetv2 (IncResv2)~\cite{szegedy2017inception}, ResNetv2-101 (Resv2-101)~\cite{he2016identity}, ResNetv2-152 (Resv2-152)~\cite{he2016identity}, PNASNet~\cite{liu2018progressive} and NASNet~\cite{zoph2018learning}), three adversarially trained models (\ie, Ens3Inceptionv3 (Ens3-Incv3), Ens4Inceptionv3 (Ens4-Incv3) and EnsInceptionResNetv2 (Ens-IncResv2)~\cite{tramer2017ensemble}). All models are publicly available\footnote{\url{https://github.com/tensorflow/models/tree/master/research/slim}}\textsuperscript{,} \footref{note1}. In addition, another eight stronger defense models are also used to evaluate the generated adversarial examples, including HGD~\cite{liao2018defense}, R\&P~\cite{xie2018mitigating}, NIPS-r3\footnote{\url{https://github.com/anlthms/nips-2017/tree/master/mmd}}, Bit-Red~\cite{xu2018feature}, JPEG~\cite{guo2018countering}, FD~\cite{liu2019feature}, ComDefend~\cite{jia2019comdefend} and RS~\cite{jia2020certified}.

\subsection{Baselines}
\label{sec:setup:baseline}
  We utilize several input-transformation-based black-box attack methods (\eg, DIM~\cite{xie2019improving}, TIM~\cite{dong2019evading}, SIM~\cite{lin2020nesterov}, CIM~\cite{yang2021adversarial}, Admix~\cite{wang2021admix}) to compare with our method. By default, we incorporate these methods into MIFGSM~\cite{dong2018boosting}, \ie, using the formula of \cref{equ:T}. Besides, we also combine these input-transformation-based methods together to form the strongest baseline, called Admix-DI-SI-CI-MIFGSM (ADSCM). We also compare our AITL with recently proposed AutoMA~\cite{yuan2021automa}.

  In addition, we use a random selection method instead of the AITL model to choose the combination of image transformations used in the attack, which is denoted as \verb|Random| method. From the experiments in \cref{sec:exper:analysis}, we know that the geometry-based image transformations are more effective than most color-based image transformations (except \verb|Scale|) to improve the transferability of adversarial examples. So we exclude these color-based image transformations (except \verb|Scale|) from the transformation candidates, and finally choose 12 transformations in \verb|Random| method, including \verb|Admix|, \verb|Scale|, \verb|Admix-and-Scale|, \verb|Crop|, \verb|Resize|, \verb|Rotate|, \verb|ShearX|, \verb|ShearY|, \verb|TranslateX|, \verb|TranslateY|, \verb|Reshape|, \verb|Cutout|.

  For the hyperparameters used in baselines, the decay factor $\mu$ in MIFGSM~\cite{dong2018boosting} is set to 1.0. The transformation probability $p$ in DIM~\cite{xie2019improving} is set to 0.7. The kernel size $k$ in TIM~\cite{dong2019evading} is set to 7. For a fair comparison of different methods, we control the number of repetitions per iteration in all methods to 5, \ie, the number of scale copies $m$ in SIM~\cite{lin2020nesterov} is set to 5, $m_1$ and $m_2$ in Admix~\cite{wang2021admix} are set to 1 and 5, respectively, $m$ in AutoMA~\cite{yuan2021automa} is set to 5, and $N$ in our AITL is also set to 5.

\subsection{Image Transformation Operations}
\label{sec:setup:operation}
  Inspired by~\cite{cubuk2019autoaugment,cubuk2020randaugment} and considering the characteristic in the task of adversarial attack, we totally select 20 image transformation operations as candidates, including \verb|Admix|, \verb|Scale|, \verb|Admix-and-Scale|, \verb|Brightness|, \verb|Color|, \verb|Contrast|, \verb|Sharpness|, \verb|Invert|, \verb|Hue|, \verb|Saturation|, \verb|Gamma|, \verb|Crop|, \verb|Resize|, \verb|Rotate|, \verb|ShearX|, \verb|ShearY|, \verb|TranslateX|, \verb|TranslateY|, \verb|Reshape|, \verb|Cutout|.
  In this section, we introduce each transformation operation in detail, and give the range of magnitude towards each operation.

\subsubsection{Geometry-based Operations.}
  Many image transformation operations are based on affine transformation. Assuming that the position of a certain pixel in the image is $(x, y)$, the operation of affine transformation can be formulated as:
  \newcommand{\block}[1]{
    \underbrace{\begin{matrix}1 & \cdots & 1\end{matrix}}_{#1}
  }
  \begin{equation}
    \begin{pmatrix}
      x' \\ y' \\ 1
    \end{pmatrix}
    =
    \underbrace{
      \begin{pmatrix}
      a_{11} & a_{12} & a_{13} \\
      a_{21} & a_{22} & a_{23} \\
      0 & 0 & 1 \\
      \end{pmatrix}
    }_{A}
    \begin{pmatrix}
      x \\ y \\ 1
    \end{pmatrix},
  \end{equation}
  where $A$ is the affine matrix, and $(x',y')$ is the position of the pixel after transformation.
  \begin{itemize}
    \item \textbf{Rotation.} The affine matrix in the \verb|Rotation| operation is:
    \begin{equation}
      \begin{pmatrix}
        \cos \theta & -\sin \theta & 0 \\
        \sin \theta & \cos \theta & 0 \\
        0 & 0 & 1
      \end{pmatrix},
    \end{equation}
    where $\theta$ is the angle of rotation. In implementation, we set $\theta \in [-30 ^\circ, 30^\circ]$.
    \item \textbf{ShearX.} The operation of \verb|ShearX| is used to shear the image along $x$-axis, whose affine matrix is:
    \begin{equation}
      \begin{pmatrix}
        1 & a & 0 \\
        0 & 1 & 0 \\
        0 & 0 & 1
      \end{pmatrix},
    \end{equation}
    where $a$ is used to control the magnitude of shearing. In implementation, we set $a \in [-0.5, 0.5]$.
    \item \textbf{ShearY.} The operation of \verb|ShearY| is used to shear the image along $y$-axis, whose affine matrix is:
    \begin{equation}
      \begin{pmatrix}
        1 & 0 & 0 \\
        a & 1 & 0 \\
        0 & 0 & 1
      \end{pmatrix},
    \end{equation}
    where $a$ is used to control the magnitude of shearing. In implementation, we set $a \in [-0.5, 0.5]$.
    \item \textbf{TranslateX.} The operation of \verb|TranslateX| is used to translate the image along $x$-axis, whose affine matrix is:
    \begin{equation}
      \begin{pmatrix}
        1 & 0 & a \\
        0 & 1 & 0 \\
        0 & 0 & 1
      \end{pmatrix},
    \end{equation}
    where $a$ is used to control the magnitude of translating. In implementation, we set $a \in [-0.4, 0.4]$.
    \item \textbf{TranslateY.} The operation of \verb|TranslateY| is used to translate the image along $y$-axis, whose affine matrix is:
    \begin{equation}
      \begin{pmatrix}
        1 & 0 & 0 \\
        0 & 1 & a \\
        0 & 0 & 1
      \end{pmatrix},
    \end{equation}
    where $a$ is used to control the magnitude of translating. In implementation, we set $a \in [-0.4, 0.4]$.
    \item \textbf{Reshape.} We use the operation of \verb|Reshape| to represent any affine transformation, which has a complete 6 degrees of freedom. In implementation, we set $a_{11}, a_{22} \in [0.5, 1.5]$ and $a_{12}, a_{13}, a_{21}, a_{23} \in [-0.5, 0.5]$.
    \item \textbf{Resizing.} For the operation of \verb|Resizing|, the original image with the size of $299 \times 299 \times 3$ is first randomly resized to $h \times w \times 3$, where $h, w \in [299, 330]$, and then zero padded to the size of $330 \times 330 \times 3$. Finally, the image is resized back to the size of $299 \times 299 \times 3$. 
    \item \textbf{Crop.} The operation of \verb|Crop| randomly crops a region of $h \times w \times 3$ size from the original image, where $h, w \in [279, 299]$, and then resizes the cropped region to the size of $299 \times 299 \times 3$.
  \end{itemize}
\subsubsection{Color-based Operations.}
  \begin{itemize}
    \item \textbf{Brightness.} The operation of \verb|Brightness| is used to randomly adjust the brightness of the image with a parameter $\alpha \in [0.5, 1.5]$ to control the magnitude. $\alpha = 0$ refers to a black image and $\alpha=1$ refers to the original image.
    \item \textbf{Color.} The operation of \verb|Color| is used to randomly adjust the color balance of the image with a parameter $\alpha \in [0.5, 1.5]$ to control the magnitude. $\alpha = 0$ refers to a black-and-white image and $\alpha=1$ refers to the original image.
    \item \textbf{Contrast.} The operation of \verb|Contrast| is used to randomly adjust the contrast of the image with a parameter $\alpha \in [0.5, 1.5]$ to control the magnitude. $\alpha = 0$ refers to a gray image and $\alpha=1$ refers to the original image.
    \item \textbf{Sharpness.} The operation of \verb|Sharpness| is used to randomly adjust the sharpness of the image with a parameter $\alpha \in [0.5, 1.5]$ to control the magnitude. $\alpha = 0$ refers to a blurred image and $\alpha=1$ refers to the original image.
    \item \textbf{Hue.} The operation of \verb|Hue| is used to randomly adjust the hue of the image with a parameter $\alpha \in [-0.2, 0.2]$ to control the magnitude. The image is first converted from RGB color space to HSV color space. After adjusting the image in the H channel, the processed image is then converted back to RGB color space.
    \item \textbf{Saturation.} The operation of \verb|Saturation| is used to randomly adjust the saturation of the image with a parameter $\alpha \in [0.5, 1.5]$ to control the magnitude. The image is first converted from RGB color space to HSV color space. After adjusting the image in the S channel, the processed image is then converted back to RGB color space.
    \item \textbf{Gamma.} The operation of \verb|Gamma| performs the gamma transformation on the image with a parameter $\alpha \in [0.6, 1.4]$ to control the magnitude.
    \item \textbf{Invert.} The operation of \verb|Invert| inverts the pixels of the image, \eg changing the value of pixel from 0 to 255 and changing the value of pixel from 255 to 0.
    \item \textbf{Scale.} The operation of \verb|Scale| is borrowed from SIM~\cite{lin2020nesterov}, which scales the original image $x$ with a parameter $m \in [0, 4]$:
    \begin{equation}
      \tilde{x} = \frac{x}{2^m}.
    \end{equation}
  \end{itemize}
\subsubsection{Other Operations.}
  \begin{itemize}
    \item \textbf{Admix.} The operation of \verb|Admix| is borrowed from Admix~\cite{wang2021admix}, which interpolates the original image $x$ with another randomly selected image $x'$ as follows:
    \begin{equation}
      \tilde{x} = x + \eta \cdot x',
    \end{equation}
    where $\eta =0.2$ is used to control the magnitude of transformation.

    \item \textbf{Admix-and-Scale.} Since the operation of \verb|Admix| changes the range of pixel values in the image, the processed image is likely to exceed the range of $[0, 255]$. So we combine the operation of \verb|Admix| and \verb|Scale| as a new operation of \verb|Admix-and-Scale|.
    \item \textbf{Cutout.} The operation of \verb|Cutout| cuts a piece of $60\times 60$ region from the original image and pads with zero.
  \end{itemize}

\subsection{The detailed structure of AITL}
\label{sec:setup:network}
  \begin{figure}[t]
    \centering
    \includegraphics[width=0.8\linewidth]{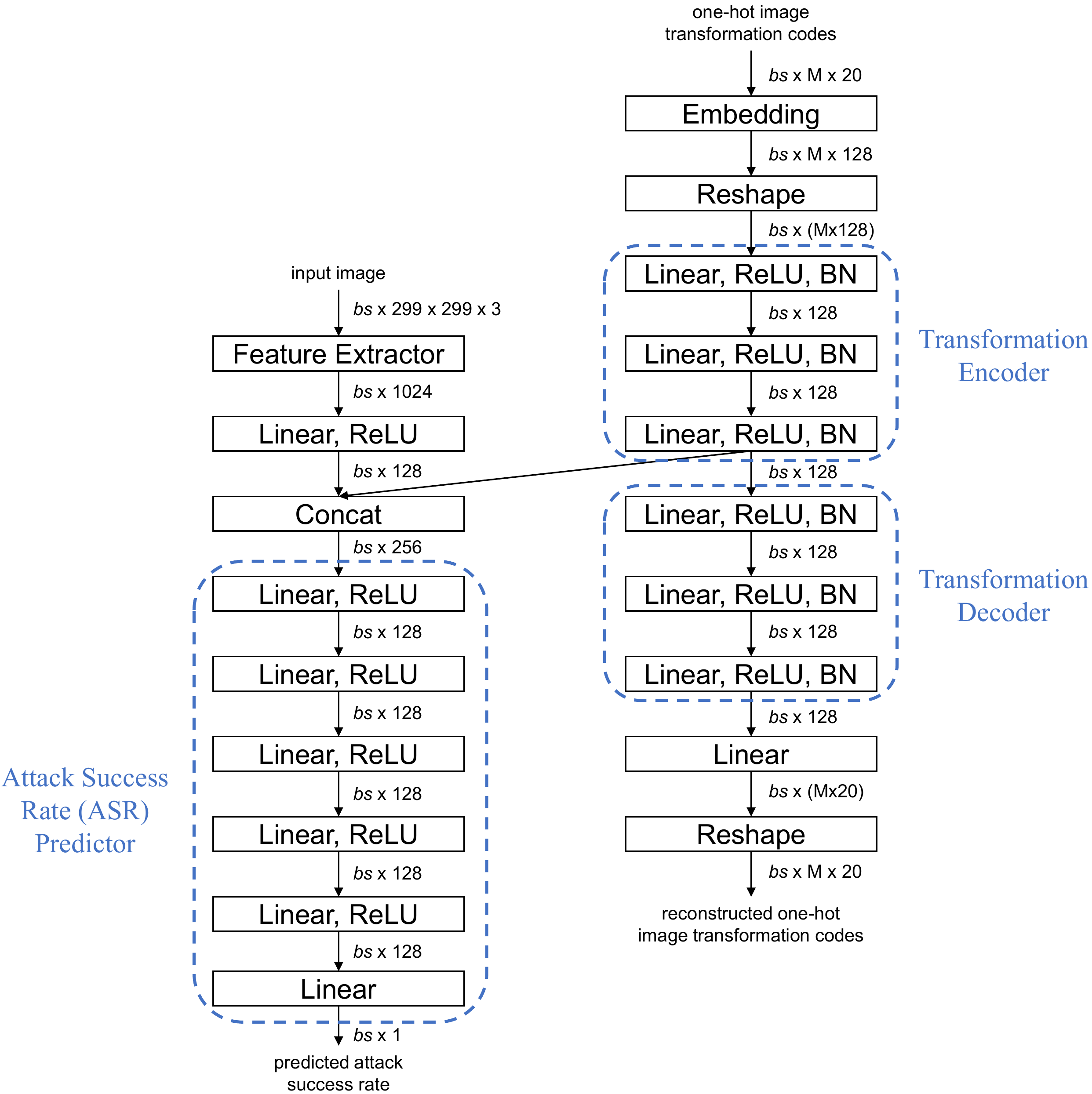}
    \caption{The specific structure of the Adaptive Image Transformation Learner. The feature extractor utilizes the pre-trained MobileNetv2-1.0~\cite{sandler2018mobilenetv2} as initialization and is further finetuned together with the whole structure. The image features are extracted from the \texttt{Global\_Pool} layer. The \textit{bs} represents the batch size and $M$ represents the number of image transformation operations in each combination}
    \label{fig:network}
  \end{figure}
  The specific structure of the AITL network is shown in \cref{fig:network}. The feature extractor utilizes the pre-trained MobileNetv2-1.0~\cite{sandler2018mobilenetv2} as initialization and is further finetuned together with the whole structure. The image features are extracted from the \texttt{Global\_Pool} layer. The \textit{bs} represents the batch size and $M$ represents the number of image transformation operations in each combination.

\section{Discussion}
\label{sec:discussion}
  The works most relevant to ours are AutoMA~\cite{yuan2021automa} and ATTA~\cite{wu2021improving}. In this section, we give a brief discussion on the differences between our work and theirs.

\subsection{Comparison with AutoMA.}
  Both AutoMA~\cite{yuan2021automa} and our AITL utilize a trained model to select the suitable image transformation for each image to improve the transferability of generated adversarial examples. The differences between the two works lie in the following aspects: 1) AutoMA adopts a reinforcement learning framework to search for a strong augmentation policy. Since the reward function is non-differentiable, Proximal Policy Optimization algorithm~\cite{schulman2017proximal} as a trade-off solution, is utilized to update the model. Differently, we design an end-to-end differentiable model, which makes the optimization process easier to converge, leading to better image transformation combinations for various images. 2) AutoMA only takes the single image transformation as augmentation, while we further consider serialized combinations of different image transformations, so that more diverse image transformation choices can be taken during the generation of adversarial examples. 3) More kinds of image transformations are considered in the candidate set in our method (20 in our AITL \vs~10 in AutoMA). Extensive experiments in \cref{sec:exper:attack} show that our method significantly outperforms AutoMA, which demonstrates the superiority of our AITL.

\subsection{Comparison with ATTA.}
  Both ATTA~\cite{wu2021improving} and our AITL use transformed images to generate the adversarial examples, but there are also significant differences between the two. ATTA directly models the image transformation process through a pixel-to-pixel network to generate the transformed image. Since the solution space is quite large ($H*W*C$), it may be difficult to approach the optimal solution. On the contrary, our AITL turns to seek the optimal combinations of various existing transformations. The solution space of our method is reduced sharply to the number of image transformation operations ($\sim$20), so the optimization difficulty is well controlled and thus more tractable. The detailed experimental comparison and analysis between ATTA and our AITL are provided in \cref{sec:add:atta}.

\section{Additional Experiments}
\label{sec:extra}
\subsection{Ablation Study}
  In this section, we analyze of effects of the number $N$ of repetitions of the image transformation combination used in each attack step and the number $M$ of image transformation operations in each combination.

\begin{table*}[!htbp]
  \centering
  \caption{The attack success rates under different number of repetitions of the image transformation combination during each attack step, \ie, $N$ in \cref{fig:comparison}. The adversarial examples are crafted on Incv3. $^*$ indicates the white-box model}
  \begin{subtable}{0.75\textwidth}
    \centering
    \caption{The evaluation against 7 normally trained models}
    \resizebox{\textwidth}{!}{
      \begin{tabular}{c|ccccccc}
        \hline
        \noalign{\smallskip}
        & Incv3$^*$ & Incv4 & IncResv2 & Resv2-101 & Resv2-152 & PNASNet & NASNet \\
        \noalign{\smallskip}
        \hline
        \noalign{\smallskip}
        AITL ($N$=5) & 99.8& 95.8& 94.1& 88.8& 90.1& 94.1& 94.0 \\
        AITL ($N$=10) & 100& 98.0& 97.1& 92.5& 93.1& 96.2& 96.9\\
        AITL ($N$=15) & 99.9& 98.5& 98.2& 94.3& 94.4& 97.1& 97.0 \\
        \noalign{\smallskip}
        \hline
      \end{tabular}
    }
  \end{subtable}
  \begin{subtable}{\textwidth}
    \caption{The evaluation against 11 defense models}
    \resizebox{\textwidth}{!}{
      \begin{tabular}{c|ccccccccccc}
        \hline
        \noalign{\smallskip}
        & Incv3$_{\texttt{ens3}}$ & Incv3$_{\texttt{ens4}}$ & IncResv2$_{\texttt{ens}}$ & HGD & R\&P & NIPS-r3 & Bit-Red & JPEG & FD & ComDefend & RS \\
        \noalign{\smallskip}
        \hline
        \noalign{\smallskip}
        AITL ($N$=5) & 69.9& 65.8& 43.4& 50.4& 46.9& 59.9& 51.6& 87.1& 73.0& 83.2& 39.5\\
        AITL ($N$=10) & 75.7& 73.5& 50.7& 60.5& 55.8& 69.5& 57.7& 93.8& 78.2& 88.5& 46.6 \\
        AITL ($N$=15) & 80.4& 78.0& 54.8& 66.1& 58.6& 72.0& 60.8& 95.0& 80.8& 91.0& 47.3\\
        \noalign{\smallskip}
        \hline
      \end{tabular}
    }
  \end{subtable}
  \label{tab:N}
\end{table*}

\subsubsection{The effect of the number $N$.}
  In order to alleviate the impact of the instability caused by the randomness in image transformation on the generated adversarial examples, many existing methods~\cite{lin2020nesterov,wang2021admix,yang2021adversarial} repeat the image transformation many times in parallel during each step of the attack. We change the number $N$ of repetitions of the image transformation combination used in each attack step, ranging from 5 to 15. As shown in \cref{tab:N}, more repetitions mean higher attack success rates, but it also brings about a higher computation cost. By increasing $N$ from 5 to 10, the improvements of attack success rates on both normally trained models and defense models are significant. When further increasing $N$ to 15, the improvement is slightly smaller, especially on the normally trained models.
  For a fair comparison of different methods, we control the number of repetitions per iteration in all methods to 5, \ie, $m=5$ in SIM~\cite{lin2020nesterov}, $m_2=5$ in Admix~\cite{wang2021admix}, $m=5$ in AutoMA~\cite{yuan2021automa} and $N=5$ in our AITL.

\begin{table*}
  \centering
  \caption{The attack success rates under different number of image transformation operations in each combination, \ie, $M$ in \cref{fig:training} and \cref{fig:inference}. The adversarial examples are crafted on Incv3. $^*$ indicates the white-box model}
  \begin{subtable}{0.75\textwidth}
    \centering
    \caption{The evaluation against 7 normally trained models}
    \resizebox{\textwidth}{!}{
      \begin{tabular}{c|ccccccc}
        \hline
        \noalign{\smallskip}
        & Incv3$^*$ & Incv4 & IncResv2 & Resv2-101 & Resv2-152 & PNASNet & NASNet \\
        \noalign{\smallskip}
        \hline
        \noalign{\smallskip}
        AITL ($M$=2) & 99.3& 93.6& 91.5& 87.2& 87.4& 91.8& 91.9\\
        AITL ($M$=3) & 99.2& 94.5& 92.6& 88.1& 88.7& 93.0& 93.0\\
        AITL ($M$=4) & 99.8& 95.8& 94.1& 88.8& 90.1& 94.1& 94.0\\
        \noalign{\smallskip}
        \hline
      \end{tabular}
    }
  \end{subtable}
  \begin{subtable}{\textwidth}
    \caption{The evaluation against 11 defense models}
    \resizebox{\textwidth}{!}{
      \begin{tabular}{c|ccccccccccc}
        \hline
        \noalign{\smallskip}
        & Incv3$_{\texttt{ens3}}$ & Incv3$_{\texttt{ens4}}$ & IncResv2$_{\texttt{ens}}$ & HGD & R\&P & NIPS-r3 & Bit-Red & JPEG & FD & ComDefend & RS \\
        \noalign{\smallskip}
        \hline
        \noalign{\smallskip}
        AITL ($M$=2) & 64.6& 58.6& 36.0& 42.5& 38.9& 52.1& 41.7& 84.9& 65.0& 74.9& 32.0\\
        AITL ($M$=3) & 67.8& 63.8& 38.9& 48.2& 43.4& 56.6& 45.7& 86.0& 67.4& 80.6& 36.8\\
        AITL ($M$=4) & 69.9& 65.8& 43.4& 50.4& 46.9& 59.9& 51.6& 87.1& 73.0& 83.2& 39.5\\
        \noalign{\smallskip}
        \hline
      \end{tabular}
    }
  \end{subtable}
  \label{tab:M}
\end{table*}

\subsubsection{The effect of the number $M$.}
  We also investigate the effect of the number $M$ of image transformation operations in each combination on the attack success rates. The experimental results are shown in \cref{tab:M}. We find that increasing $M$ from 2 to 3 can bring an obvious improvement in the attack success rates, but the improvement is marginal when further increasing $M$ to 4. Therefore, considering the computational efficiency, we set $M$ to 4 in other experiments instead of further increasing $M$. Noting that the number of transformations used in the strongest baseline ADSCM is also 4 (\eg, Admix, resize, crop and scale), which is the same as ours, formulating fair experimental comparisons.

\subsection{More Results of Attack on the Single Model}
\label{sec:add:single}
  We conduct the adversarial attack on more models under the setting of the single model in this section. We choose Incv4, IncResv2 and Resv2-101 as the white-box model to conduct the attack, respectively, and evaluate the generated adversarial examples against both normally trained models and defense models. The results are shown in \cref{tab:single:incv4}, \cref{tab:single:incresv2} and \cref{tab:single:resv2}, respectively. From the results we can clearly conclude that, when choosing different models as the white-box model to conduct the attack, our proposed AITL consistently achieve higher attack success rates on various black-box models, especially on the defense models. It also shows that the optimal combinations of image transformations selected by the well-trained AITL specific to each image can successfully attack different models, \ie, our AITL has a good generalization.

\begin{table*}[!htbp]
  \centering
  \caption{Attack success rates (\%) of adversarial attacks against 7 normally trained models and 11 defense models under \textbf{single model} setting. The adversarial examples are crafted on \textbf{Incv4}. $^*$ indicates the white-box model. $^\dagger$ The results of AutoMA~\cite{yuan2021automa} are cited from their original paper}
  \begin{subtable}{\textwidth}
    \centering
    \caption{The evaluation against 7 normally trained models}
    \resizebox{0.75\textwidth}{!}{
      \begin{tabular}{c|ccccccc}
        \hline\noalign{\smallskip}
        & Incv3 & Incv4$^*$ & IncResv2 & Resv2-101 & Resv2-152 & PNASNet & NASNet \\
        \noalign{\smallskip}
        \hline
        \noalign{\smallskip}
        MIFGSM~\cite{dong2018boosting} & 65.9& \textbf{100}& 55.0& 37.3& 38.0& 48.6& 46.6\\
        DIM~\cite{xie2019improving} & 84.9& 99.5& 79.4& 57.9& 56.3& 68.4& 67.3\\
        SIM~\cite{lin2020nesterov} & 87.2& 99.9& 81.9& 69.4& 69.0&72.8& 75.4\\
        CIM~\cite{yang2021adversarial} & 90.8& 99.9& 84.3& 57.1& 56.7& 67.1& 68.0\\
        Admix~\cite{wang2021admix} & 81.9& 99.9& 76.3& 63.9& 61.4& 68.3& 69.6\\
        ADSCM & 92.6& 99.8& 89.5& 77.1& 78.2& 82.2& 82.7\\
        Random & 95.4 & 99.9& 93.0& 77.7& 78.0& 84.4&84.9\\
        AITL (ours) & \textbf{97.0}& 99.8& \textbf{95.3}& \textbf{87.8}& \textbf{88.9}& \textbf{92.4}& \textbf{93.7}\\
        \noalign{\smallskip}
        \hline
        \noalign{\smallskip}
        AutoMA-TIM$^\dagger$~\cite{wu2021improving} & 86.8& 98.1& 78.8& 71.4&- &- &- \\
        AITL-TIM (ours) &\textbf{94.7} &\textbf{99.8} &\textbf{92.8} &\textbf{89.4} &\textbf{89.4} &\textbf{91.7} &\textbf{92.4}\\
        \noalign{\smallskip}
        \hline
      \end{tabular}
    }
  \end{subtable}
  \begin{subtable}{\textwidth}
    \caption{The evaluation against 11 defense models}
    \resizebox{\textwidth}{!}{
      \begin{tabular}{c|ccccccccccc}
        \hline
        \noalign{\smallskip}
        & Incv3$_{\texttt{ens3}}$ & Incv3$_{\texttt{ens4}}$ & IncResv2$_{\texttt{ens}}$ & HGD & R\&P & NIPS-r3 & Bit-Red & JPEG & FD & ComDefend & RS \\
        \noalign{\smallskip}
        \hline
        \noalign{\smallskip}
        MIFGSM~\cite{dong2018boosting} & 20.1& 18.6& 9.4& 9.3& 9.0& 13.7& 21.0& 37.5& 38.4& 29.7& 17.2\\
        DIM~\cite{xie2019improving} & 36.1& 34.1& 18.9& 25.9& 21.6& 30.7& 28.1& 60.8& 47.8& 49.0& 23.8\\
        SIM~\cite{lin2020nesterov} & 55.0& 50.8& 32.9& 35.5& 33.4& 42.1& 38.7& 71.8& 57.8& 62.5& 32.1\\
        CIM~\cite{yang2021adversarial} & 37.5& 33.9& 19.9& 29.9& 21.2& 30.4& 29.0& 62.6& 46.8& 47.4& 23.3\\
        Admix~\cite{wang2021admix} & 41.4& 39.2& 24.8& 22.8& 23.8& 31.4& 34.0& 61.2& 53.6& 53.1& 27.9\\
        ADSCM & 62.1& 57.6& 39.7& 45.7& 43.7& 53.0& 47.0& 80.6& 66.8& 73.0& 37.6\\
        Random & 56.8& 50.1& 33.8& 40.3& 37.1& 48.0& 40.9& 81.4& 59.8& 71.7& 32.8\\
        AITL (ours) & \textbf{75.4}& \textbf{73.7}& \textbf{55.8}& \textbf{66.6}& \textbf{59.0}& \textbf{71.4}& \textbf{56.9}& \textbf{89.5}& \textbf{76.6}& \textbf{85.1}& \textbf{46.6}\\
        \noalign{\smallskip}
        \hline
        \noalign{\smallskip}
        AutoMA-TIM$^\dagger$~\cite{wu2021improving} & 76.0 & 75.5& 67.4& 69.6& 68.0& 71.0& - & -& 82.1& -& -\\
        AITL-TIM (ours) & \textbf{82.7} & \textbf{79.6}& \textbf{72.4}& \textbf{79.6}& \textbf{71.8}& \textbf{79.4}& \textbf{62.4}& \textbf{88.3}& \textbf{83.2}& \textbf{87.2}& \textbf{57.3}\\
        \noalign{\smallskip}
        \hline
      \end{tabular}
    }
  \end{subtable}
  \label{tab:single:incv4}
\end{table*}

\begin{table*}[!htbp]
  \centering
  \caption{Attack success rates (\%) of adversarial attacks against 7 normally trained models and 11 defense models under \textbf{single model} setting. The adversarial examples are crafted on \textbf{IncResv2}. $^*$ indicates the white-box model. $^\dagger$ The results of AutoMA~\cite{yuan2021automa} are cited from their original paper}
  \begin{subtable}{\textwidth}
    \centering
    \caption{The evaluation against 7 normally trained models}
    \resizebox{0.75\textwidth}{!}{
      \begin{tabular}{c|ccccccc}
        \hline\noalign{\smallskip}
        & Incv3 & Incv4 & IncResv2$^*$ & Resv2-101 & Resv2-152 & PNASNet & NASNet \\
        \noalign{\smallskip}
        \hline
        \noalign{\smallskip}
        MIFGSM~\cite{dong2018boosting} & 72.2& 63.4& 99.3& 47.4& 46.5& 51.9& 52.0\\
        DIM~\cite{xie2019improving} & 85.3& 82.7& 98.5& 66.8& 67.2& 68.0& 72.3\\
        SIM~\cite{lin2020nesterov} & 92.0& 88.2& \textbf{99.8}& 77.9& 77.8& 77.1& 81.0\\
        CIM~\cite{yang2021adversarial} & 90.6& 88.4& 99.4& 69.4& 68.3& 70.8& 73.9\\
        Admix~\cite{wang2021admix} & 87.0& 81.4& \textbf{99.8}& 70.7& 71.5& 72.6& 75.4\\
        ADSCM & 94.0& 91.3& 99.6& 85.0& 84.4& 84.7& 86.9\\
        Random & 94.8 & 91.3& 99.2& 84.9& 84.5& 86.3& 87.8\\
        AITL (ours) & \textbf{95.7}& \textbf{94.3}& 99.1& \textbf{89.6}& \textbf{89.6}& \textbf{92.0}& \textbf{92.3}\\
        \noalign{\smallskip}
        \hline
        \noalign{\smallskip}
        AutoMA-TIM$^\dagger$~\cite{wu2021improving} & 87.2& 85.6& 98.3& 79.7&- &- &- \\
        AITL-TIM (ours) &\textbf{94.3} &\textbf{92.1} &\textbf{99.0} &\textbf{91.6} &\textbf{91.3} &\textbf{93.3} &\textbf{95.2}\\
        \noalign{\smallskip}
        \hline
      \end{tabular}
    }
  \end{subtable}
  \begin{subtable}{\textwidth}
    \caption{The evaluation against 11 defense models}
    \resizebox{\textwidth}{!}{
      \begin{tabular}{c|ccccccccccc}
        \hline
        \noalign{\smallskip}
        & Incv3$_{\texttt{ens3}}$ & Incv3$_{\texttt{ens4}}$ & IncResv2$_{\texttt{ens}}$ & HGD & R\&P & NIPS-r3 & Bit-Red & JPEG & FD & ComDefend & RS \\
        \noalign{\smallskip}
        \hline
        \noalign{\smallskip}
        MIFGSM~\cite{dong2018boosting} & 28.8& 23.0& 15.0& 17.1& 14.2& 18.1& 25.0& 48.5& 43.9& 38.7& 18.8\\
        DIM~\cite{xie2019improving} & 49.3& 43.3& 30.1& 39.8& 32.0& 42.9& 33.5& 69.4& 54.9& 60.3& 26.3\\
        SIM~\cite{lin2020nesterov} & 63.3& 55.1& 48.5& 47.3& 41.2& 51.8& 43.8& 80.6& 62.5& 70.9& 33.3\\
        CIM~\cite{yang2021adversarial} & 53.8& 45.4& 32.5& 44.6& 35.0& 45.4& 33.5& 73.5& 54.6& 58.3& 26.1\\
        Admix~\cite{wang2021admix} & 51.6& 44.3& 35.1& 32.2& 30.2& 39.9& 39.6& 70.5& 58.1& 61.7& 30.1\\
        ADSCM & 70.4& 65.4& 51.7& 56.9& 55.1& 63.1& 53.8& 83.1& 72.3& 77.5& 40.1\\
        Random & 67.5& 59.5& 48.6& 54.1& 49.6& 62.5& 47.6& 86.6& 67.4& 79.8& 36.0\\
        AITL (ours) & \textbf{79.9}& \textbf{74.6}& \textbf{65.4}& \textbf{71.5}& \textbf{67.5}& \textbf{75.8}& \textbf{62.3}& \textbf{89.8}& \textbf{79.8}& \textbf{86.1}& \textbf{49.4}\\
        \noalign{\smallskip}
        \hline
        \noalign{\smallskip}
        AutoMA-TIM$^\dagger$~\cite{wu2021improving} & 84.2 & 82.9& 82.8& 79.9& 81.6& 83.7& - & -& 87.2& -& -\\
        AITL-TIM (ours) & \textbf{88.6} & \textbf{85.0}& \textbf{85.4}& \textbf{83.4}& \textbf{84.2}& \textbf{86.2}& \textbf{71.7}& \textbf{89.4}& \textbf{88.4}& \textbf{90.4}& \textbf{65.4}\\
        \noalign{\smallskip}
        \hline
      \end{tabular}
    }
  \end{subtable}
  \label{tab:single:incresv2}
\end{table*}

\begin{table*}[!htbp]
  \centering
  \caption{Attack success rates (\%) of adversarial attacks against 7 normally trained models and 11 defense models under \textbf{single model} setting. The adversarial examples are crafted on \textbf{Resv2-101}. $^*$ indicates the white-box model. $^\dagger$ The results of AutoMA~\cite{yuan2021automa} are cited from their original paper}
  \begin{subtable}{\textwidth}
    \centering
    \caption{The evaluation against 7 normally trained models}
    \resizebox{0.75\textwidth}{!}{
      \begin{tabular}{c|ccccccc}
        \hline\noalign{\smallskip}
        & Incv3 & Incv4 & IncResv2 & Resv2-101$^*$ & Resv2-152 & PNASNet & NASNet \\
        \noalign{\smallskip}
        \hline
        \noalign{\smallskip}
        MIFGSM~\cite{dong2018boosting} & 50.5& 40.6& 41.9& 98.9& 86.5& 61.4& 60.6\\
        DIM~\cite{xie2019improving} & 67.3& 58.0& 61.2& 98.6& 92.5& 75.0& 77.6\\
        SIM~\cite{lin2020nesterov} & 59.4& 49.8& 53.4& 99.5& 93.8& 75.3& 76.4\\
        CIM~\cite{yang2021adversarial} & 77.9& 70.2& 73.2& 98.9& 95.3& 81.4& 82.4\\
        Admix~\cite{wang2021admix} & 54.2& 45.4& 46.3& \textbf{99.7}& 90.2& 71.7& 70.4\\
        ADSCM & 73.9& 63.3& 68.5& 99.5& 94.9& 83.3& 84.0\\
        Random & 80.2 & 74.3& 78.5& 99.1& 94.3& 88.3& 89.5\\
        AITL (ours) & \textbf{81.5}& \textbf{78.3}& \textbf{79.4}& 99.3& \textbf{96.1}& \textbf{91.2}& \textbf{91.3}\\
        \noalign{\smallskip}
        \hline
        \noalign{\smallskip}
        AutoMA-TIM$^\dagger$~\cite{wu2021improving} & 75.0& 71.2& 69.3& 97.0&- &- &- \\
        AITL-TIM (ours) &\textbf{79.5} &\textbf{75.4} &\textbf{73.8} &\textbf{98.3} &\textbf{96.2} &\textbf{89.2} &\textbf{89.5}\\
        \noalign{\smallskip}
        \hline
      \end{tabular}
    }
  \end{subtable}
  \begin{subtable}{\textwidth}
    \caption{The evaluation against 11 defense models}
    \resizebox{\textwidth}{!}{
      \begin{tabular}{c|ccccccccccc}
        \hline
        \noalign{\smallskip}
        & Incv3$_{\texttt{ens3}}$ & Incv3$_{\texttt{ens4}}$ & IncResv2$_{\texttt{ens}}$ & HGD & R\&P & NIPS-r3 & Bit-Red & JPEG & FD & ComDefend & RS \\
        \noalign{\smallskip}
        \hline
        \noalign{\smallskip}
        MIFGSM~\cite{dong2018boosting} & 34.3& 29.9& 20.1& 22.7& 20.2& 25.2& 28.6& 41.3& 47.3& 43.5& 26.0\\
        DIM~\cite{xie2019improving} & 52.6& 45.4& 33.6& 35.9& 33.4& 41.4& 37.3& 61.5& 57.9& 62.0& 35.9\\
        SIM~\cite{lin2020nesterov} & 46.2& 43.0& 29.0& 32.3& 29.6& 36.9& 37.9& 52.4& 58.3& 59.2& 36.5\\
        CIM~\cite{yang2021adversarial} & 64.7& 60.2& 44.1& 51.6& 45.9& 56.5& 44.5& 73.0& 62.7& 70.5& 42.6\\
        Admix~\cite{wang2021admix} & 37.5& 35.1& 21.9& 25.1& 21.7& 28.1& 34.0& 45.7& 53.4& 52.4& 32.2\\
        ADSCM & 60.2& 55.3& 41.1& 43.4& 41.0& 51.1& 49.0& 67.6& 66.8& 70.2& 47.5\\
        Random & 66.5& 62.7& 47.3& 52.1& 48.1& 60.1& 52.4& 75.2& 70.8& 78.6& 47.7\\
        AITL (ours) & \textbf{71.6}& \textbf{67.7} & \textbf{53.7}& \textbf{58.6}& \textbf{53.3}& \textbf{62.3}& \textbf{56.0}& \textbf{77.1}& \textbf{73.6}& \textbf{80.0}& \textbf{57.9}\\
        \noalign{\smallskip}
        \hline
        \noalign{\smallskip}
        AutoMA-TIM$^\dagger$~\cite{wu2021improving} & 73.4 & 74.6& 68.2& 67.1& 67.6& 71.7& - & -& 82.3& -& -\\
        AITL-TIM (ours) & \textbf{78.5} & \textbf{78.3}& \textbf{69.3}& \textbf{72.5}& \textbf{69.7}& \textbf{76.8}& \textbf{65.2}& \textbf{75.8}& \textbf{85.1}& \textbf{81.4}& \textbf{63.1}\\
        \noalign{\smallskip}
        \hline
      \end{tabular}
    }
  \end{subtable}
  \label{tab:single:resv2}
\end{table*}

\subsection{Attacks under Different Perturbation Budgets}
\begin{figure*}
  \centering
  \begin{subfigure}{0.49\linewidth}
    \includegraphics[width=1.0\linewidth]{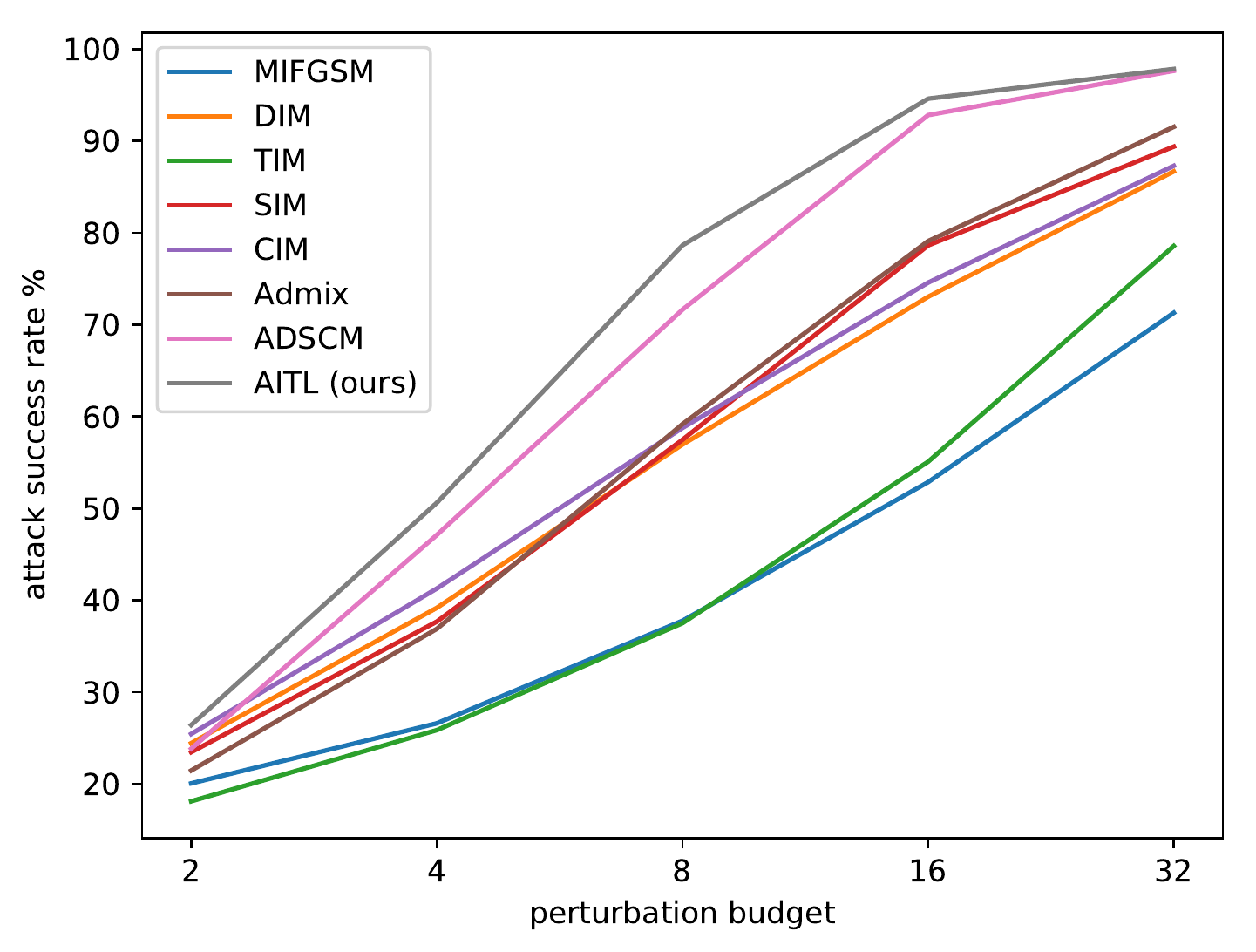}
    \caption{The average attack success rates against 7 normally trained models}
  \end{subfigure}
  \hfill
  \begin{subfigure}{0.49\linewidth}
    \includegraphics[width=1.0\linewidth]{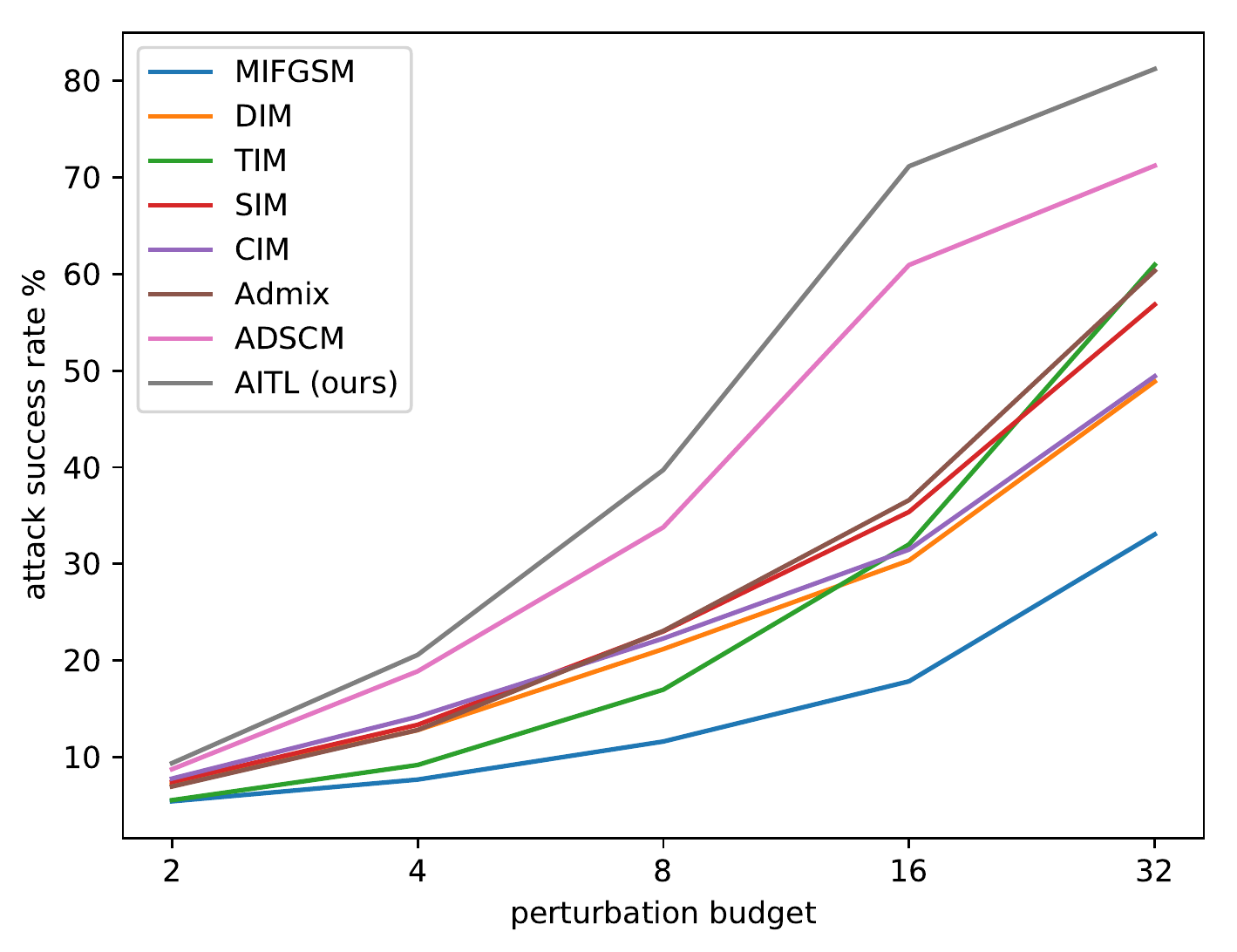}
    \caption{The average attack success rates against 11 defense models}
  \end{subfigure}
  \caption{The curve of the attack success rates \vs~different perturbation budgets}
  \label{fig:curve}
\end{figure*}

  We conduct the adversarial attacks under different perturbation budgets $\epsilon$, ranging from 2 to 32. All experiments utilize Inceptionv3~\cite{szegedy2016rethinking} model as the white-box model. The curve of the attack success rates \vs~different perturbation budgets is shown in \cref{fig:curve}. From the figure, we can clearly see that our AITL has the highest attack success rates under various perturbation budgets. Especially in the evaluation against the defense models, our AITL has an advantage of about 10\% higher attack success rates over the strongest baseline ADSCM in the case of large perturbation budgets (\eg, 16 and 32).

\subsection{Visualization}
\begin{figure}[t]
  \centering
  \includegraphics[width=1.0\linewidth]{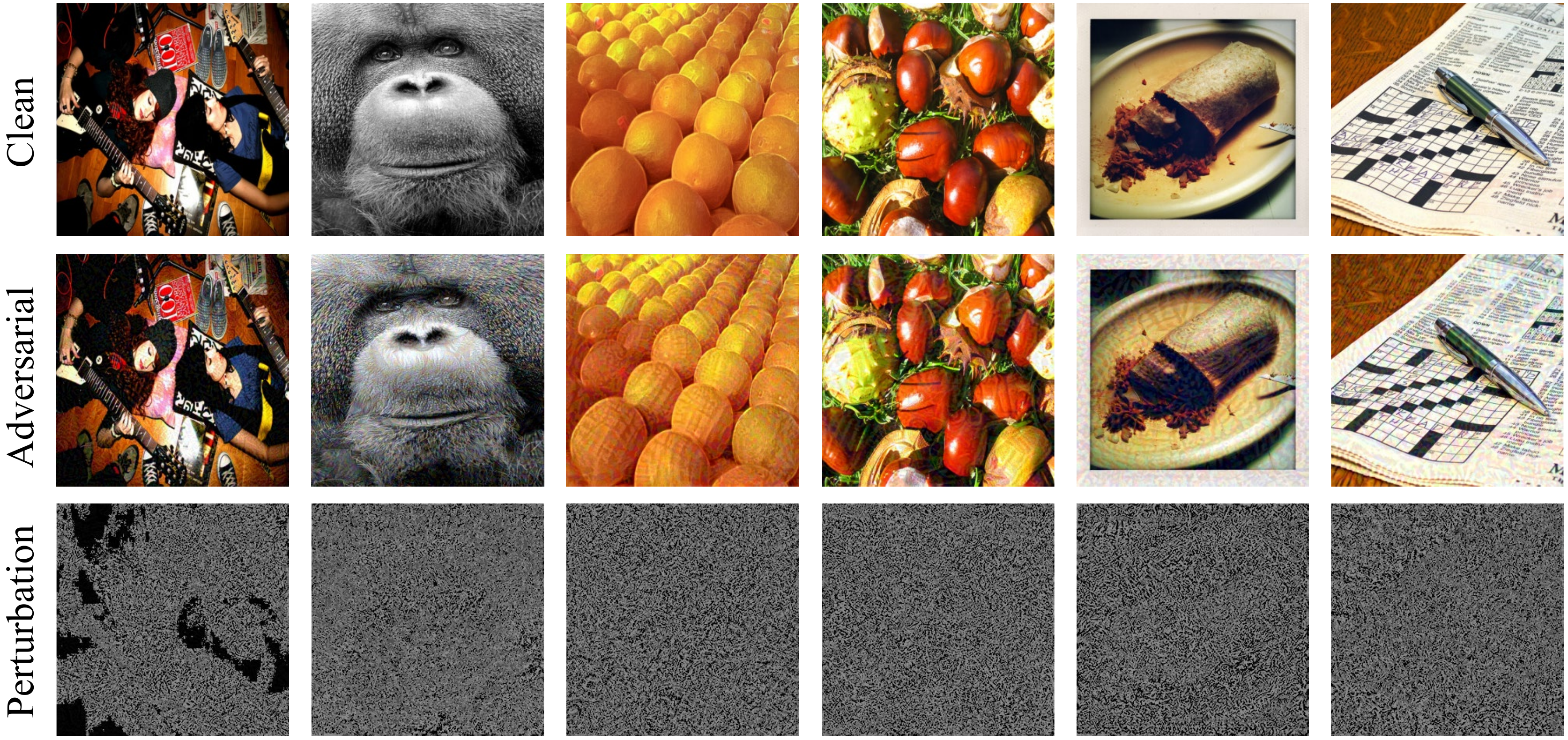}
  % \vspace{-5mm}
  \caption{Visualization of adversarial examples crafted on Incv3 by our proposed AITL}
  \label{fig:visual}
  % \vspace{-6mm}
\end{figure}

 The visualization of adversarial examples crafted on Incv3 by our proposed AITL is provided in \cref{fig:visual}.

\subsection{Combined with Other Base Attack Methods}
\label{sec:add:other}
  We combine our AITL with other base attack methods (\eg, MIFGSM~\cite{dong2018boosting}, TIM~\cite{dong2019evading}, NIM~\cite{lin2020nesterov}) and compare the results with AutoMA~\cite{yuan2021automa}. As shown in \cref{tab:other}, our method achieves significantly higher attack success rates than AutoMA~\cite{yuan2021automa} in all cases, both for normally trained and adversarially trained models, which clearly demonstrates the superiority of our AITL.
\begin{table*}[!htbp]
  \centering
  \caption{The comparison of AutoMA~\cite{yuan2021automa} and our AITL when combined with other base attack methods. The adversarial examples are crafted on Incv3. $^*$ indicates the white-box model. $^\dagger$ The results of AutoMA~\cite{yuan2021automa} are cited from their original paper}
  \resizebox{\textwidth}{!}{
    \begin{tabular}{c|ccccccc}
      \hline
      \noalign{\smallskip}
      & Incv3$^*$ & Incv4 & IncResv2 & Resv2-101 & Incv3$_{\texttt{ens3}}$ & Incv3$_{\texttt{ens4}}$ & IncResv2$_{\texttt{ens}}$ \\
      \noalign{\smallskip}
      \hline
      \noalign{\smallskip}
      MIFGSM~\cite{dong2018boosting} & \textbf{100}& 52.2 &50.6& 37.4& 15.6& 15.2& 6.4\\
      AutoMA-MIFGSM$^\dagger$~\cite{yuan2021automa} & 98.2& 91.2& 91.0& 82.5& 49.2& 49.0& 29.1\\
      AITL-MIFGSM (ours) & 99.8& \textbf{95.8}& \textbf{94.1}& \textbf{88.8}& \textbf{69.9}& \textbf{65.8}& \textbf{43.4}\\
      \noalign{\smallskip}
      \hline
      \noalign{\smallskip}
      TIM~\cite{dong2019evading} & \textbf{99.9} & 43.7 &37.6& 47.4&33.0& 30.5& 23.2\\
      AutoMA-TIM$^\dagger$~\cite{yuan2021automa} & 97.5& 80.7& 74.3& 69.3& 74.8& 74.3& 63.6\\
      AITL-TIM (ours) & 99.8& \textbf{93.4}& \textbf{92.1}& \textbf{91.9}& \textbf{81.3}& \textbf{78.9}& \textbf{69.1}\\
      \noalign{\smallskip}
      \hline
      \noalign{\smallskip}
      NIM~\cite{lin2020nesterov} & \textbf{100}& 79.0 &76.1& 64.8& 38.1& 36.6& 19.2\\
      AutoMA-NIM$^\dagger$~\cite{yuan2021automa} & 98.8& 88.4& 86.4& 80.2& 41.5& 39.3& 21.8\\
      AITL-NIM (ours) & \textbf{100}& \textbf{93.6}& \textbf{92.2}& \textbf{83.6}& \textbf{51.0}& \textbf{46.5}& \textbf{27.7}\\
      \noalign{\smallskip}
      \hline
    \end{tabular}
  }
  \label{tab:other}
\end{table*}

\subsection{Compared with Ensemble-based Attack Methods}
  We also conduct experiments to compare our AITL with several ensemble-based attack methods. Liu \etal~\cite{liu2017delving} first propose novel ensemble-based approaches to generating transferable adversarial examples. MIFGSM~\cite{dong2018boosting} studies three model fusion methods and finds that the method of fusing at the logits layer is the best. Li \etal~\cite{li2020learning} propose Ghost Networks to improve the transferability of adversarial examples by applying feature-level perturbations to an existing model to potentially create a huge set of diverse models. The experimental results of comparison between our AITL with these methods are shown in \cref{tab:ensemble}. The experimental setup is consistent with ~\cref{tab:multiple} in the manuscript, and our method achieves the best results among them.
\begin{table}[!h]
  \caption{The comparison of our AITL and other ensemble-based attack methods}
    \resizebox{\columnwidth}{!}{
        \begin{tabular}{c|cccc}
        \hline
                            & Liu \etal~\cite{liu2017delving} & Li \etal~\cite{li2020learning} & MIFGSM~\cite{dong2018boosting} & AITL(ours)     \\ \hline
        avg. of normally trained models & 74.33   & 89.76   & 82.46  & \textbf{97.36} \\
        avg. of defense models & 36.12   & 58.49   & 45.97  & \textbf{84.57} \\ \hline
        \end{tabular}
    }
    \label{tab:ensemble}
\end{table}

\subsection{Compared with ATTA and Some Clarifications}
\label{sec:add:atta}
  The comparison between our AITL and ATTA~\cite{wu2021improving} is presented in \cref{tab:atta}. The results of ATTA are directly quoted from their original paper, since the authors haven't released the code, and the details provided in the article are not enough to reproduce it. We also have tried to email the authors for more experimental details, but received no response.
  
  Here we provide some explanations and clarifications for this comparison. We have some doubts about ATTA's experimental results, since the attack success rates against advanced defense models (\eg, HGD, R\&P, NIPS-r3 and so on) is significantly higher than that against normally trained models (\eg, Incv4, IncResv2, Resv2-101), which is obviously counterintuitive. Their method seems to overfit the advanced defense models. Also, the results of some baseline methods (\eg, MIFGSM, DIM, TIM) in their paper are quite different from other works (\eg, NIM, AutoMA and ours).
  The inconsistency of experimental results may come from different experimental settings. We will do further comparisons with ATTA in the future.

\begin{table*}[!htbp]
  \centering
  \caption{The comparison between ATTA~\cite{wu2021improving} and our AITL. The adversarial examples are crafted on Incv3. $^*$ indicates the white-box model. $^\dagger$ The results of ATTA~\cite{wu2021improving} are cited from their original paper}
  \begin{subtable}{\textwidth}
    \centering
    \caption{The comparison between ATTA and AITL}
    \text{The evaluation against 7 models}
    \resizebox{0.95\textwidth}{!}{
      \begin{tabular}{c|ccccccc}
        \hline
        \noalign{\smallskip}
        & Incv3$^*$ & Incv4 & IncResv2 & Resv2-101 & Incv3$_{\texttt{ens3}}$ & Incv3$_{\texttt{ens4}}$ & IncResv2$_{\texttt{ens}}$ \\
        \noalign{\smallskip}
        \hline
        \noalign{\smallskip}
        ATTA$^\dagger$~\cite{wu2021improving} & 100& 52.9& 53.2& 44.8& 25.1& 27.9& 18.8\\
        AITL (ours) & 99.8& 95.8& 94.1& 88.8& 69.9& 65.8& 43.4\\
        \noalign{\smallskip}
        \hline
      \end{tabular}
    }
    \text{The evaluation against 6 advanced defense models}
    \resizebox{0.65\textwidth}{!}{
      \begin{tabular}{c|cccccc}
        \hline
        \noalign{\smallskip}
        & HGD & R\&P & NIPS-r3 & FD & ComDefend & RS \\
        \noalign{\smallskip}
        \hline
        \noalign{\smallskip}
        ATTA$^\dagger$~\cite{wu2021improving} & 85.9& 83.2& 89.5& 84.4& 79.9& 47.4\\
        AITL (ours) & 50.4& 46.9& 59.9& 73.0& 83.2& 39.5\\
        \noalign{\smallskip}
        \hline
      \end{tabular}
    }
  \end{subtable}
  \begin{subtable}{\textwidth}
    \centering
    \caption{The comparison between ATTA-TIM and AITL-TIM}
    \resizebox{\textwidth}{!}{
      \begin{tabular}{c|ccccccc}
        \hline
        \noalign{\smallskip}
        & Incv3$^*$ & Incv4 & IncResv2 & Resv2-101 & Incv3$_{\texttt{ens3}}$ & Incv3$_{\texttt{ens4}}$ & IncResv2$_{\texttt{ens}}$ \\
        \noalign{\smallskip}
        \hline
        \noalign{\smallskip}
        ATTA-TIM$^\dagger$~\cite{wu2021improving} & 100& 55.9& 57.1& 49.1& 27.8& 28.6& 24.9\\
        AITL-TIM (ours) & 99.8& 93.4& 92.1& 91.9& 81.3& 78.9& 69.1\\
        \noalign{\smallskip}
        \hline
      \end{tabular}
    }
  \end{subtable}
  \label{tab:atta}
\end{table*}

\section{Algorithms}
  The algorithm of training the Adaptive Image Transformation Learner is summarized in \cref{alg:training}.
  
\begin{algorithm}[t]
  \algnewcommand\algorithmicinput{\textbf{Input:}}
  \algnewcommand\Input{\item[\algorithmicinput]}
  \algnewcommand\algorithmicoutput{\textbf{Output:}}
  \algnewcommand\Output{\item[\algorithmicoutput]}

  \caption{The training of Adaptive Image Transformation Learner}
  \label{alg:training}
  \begin{algorithmic}[1]
    \Input the total training step $T$
    \Input the learning rate $\beta$
    \Input training set $(\mathcal{X}, \mathcal{Y})$, which represent the image and the corresponding label, respectively
    \Input a source classifier model $f$, $n$ target classifier models $f_1, f_2, \cdots, f_n$
    \Output Transformation Encoder $f_{en}$, Transformation Decoder $f_{de}$, ASR Predictor $f_{pre}$, Feature Extractor $f_{img}$ (denote their overall parameters as $\Theta$)
    \For {i $\in \{0, \cdots, T-1\}$ }
      \State Get an image and corresponding label $(x, y)$ from the dataset $(\mathcal{X}, \mathcal{Y})$
      \State Extract the image feature from the feature extractor $f_{img}$
      $$h_{img} = f_{img}(x)$$
      \State Randomly sample $M$ image transformation operations $t_1, t_2, \cdots, t_M$ as a combination, and represent them into one-hot codes $c_1, c_2, \cdots, c_M$
      \State Embed one-hot codes into transformation features
      $$a_1, a_2, \cdots, a_M = Embedding(c_1, c_2, \cdots, c_M) $$
      \State Concatenate the transformation features into an integrated image transformation feature vector
      $$ a = Concat(a_1, a_2, \cdots, a_M)$$
      \State Encode and decode the transformation feature vector
      \begin{gather*}
        h_{trans} = f_{en}(a) \\
        a' = f_{de}(h_{trans})
      \end{gather*}
      \State Reconstruct the image transformation operations
      $$c_1', c_2' ,\cdots, c_M' = FC(a')$$
      \State Predict the attack success rate
      \begin{gather*}
        h_{mix} = Concat(h_{trans}, h_{img}) \\
        p_{asr} = f_{pre}(h_{mix})
      \end{gather*}
      \State Generate the adversarial examples $x^{adv}$ by incorporating image transformation $t_1, t_2, \cdots, t_M$ into MIFGSM, \ie, replacing \cref{equ:g} in MIFGSM by 
      $$g_{t+1} = \mu \cdot g_t + \frac{\nabla_{x_t^{adv}} \mathcal{J}(f(t_M \circ \cdots \circ t_1(x_t^{adv})), y)}{\|\nabla_{x_t^{adv}} \mathcal{J}(f(t_M \circ \cdots \circ t_1(x_t^{adv})), y)\|_1}$$
      \algstore{myalg}
  \end{algorithmic}
\end{algorithm}
\begin{algorithm}[t]                     
  \begin{algorithmic}[1]              
      \algrestore{myalg}
      \State Calculate the corresponding actual average attack success rate $q_{asr}$ by evaluating $x^{adv}$ on $f_1, f_2, \cdots, f_n$
      $$q_{asr} = \frac{1}{n} \sum_{i=1}^n \mathbbm{1}(f_i(x^{adv}) \neq y)$$
      \State Calculate the loss function $\mathcal{L}_{total}$
      \State Update the model parameter
      $$\Theta = \Theta - \beta \cdot \nabla_{\Theta} \mathcal{L}_{total}$$
    \EndFor
  \end{algorithmic}
\end{algorithm}

  The algorithm of generating the adversarial examples with pre-trained Adaptive Image Transformation Learner is summarized in \cref{alg:inference}.
  
\begin{algorithm}[t]
  \algnewcommand\algorithmicinput{\textbf{Input:}}
  \algnewcommand\Input{\item[\algorithmicinput]}
  \algnewcommand\algorithmicoutput{\textbf{Output:}}
  \algnewcommand\Output{\item[\algorithmicoutput]}

  \caption{Generating adversarial examples with AITL}
  \label{alg:inference}
  \begin{algorithmic}[1]
    \setstretch{0.9}
    \Input the original image $x$
    \Input the number $T$ of iteration steps during attack 
    \Input the number $r$ of iterations during optimizing image transformation features
    \Input the number $N$ of repetitions of image transformation combination used in each attack step
    \Output the adversarial example $x^{adv}_T$
    \State Extract the image feature from the feature extractor $f_{img}$
    $$h_{img} = f_{img}(x)$$
    \For {i $\in \{0, \cdots, N-1\}$}
      \State Randomly sample $M$ image transformation operations $t^i_1, t^i_2, \cdots, t^i_M$ as a combination, and represent them into one-hot codes $c^i_1, c^i_2, \cdots, c^i_M$
      \State Embed one-hot codes into transformation features
      $$a^i_1, a^i_2, \cdots, a^i_M = Embedding(c^i_1, c^i_2, \cdots, c^i_M) $$
      \State Concatenate the transformation features into an integrated image transformation feature vector
      $$ a^i = Concat(a^i_1, a^i_2, \cdots, a^i_M)$$
      \State Encode the transformation feature vector
      $$ h^i_{trans} = f_{en}(a^i) $$
      \State Initialize the optimized transformation feature embedding
      $$ h^{i,0}_{trans} = h^i_{trans} $$
    \algstore{myalg}
  \end{algorithmic}
\end{algorithm}
\begin{algorithm}[!t]                     
  \begin{algorithmic}[1]              
    \algrestore{myalg}
      \For {j $\in \{0, \cdots, r-1\}$ }
        \State Predict the attack success rate
        \begin{gather*}
          h^j_{mix} = Concat(h^{i,j}_{trans}, h_{img}) \\
          p^j_{asr} = f_{pre}(h^j_{mix})
        \end{gather*}
        \State Update the transformation feature embedding
        $$ h^{i, j+1}_{trans} = h^{i,j}_{trans} + \gamma \cdot \nabla_{h^{i,j}_{trans}} p^j_{asr} $$
      \EndFor
      \State Decode the transformation feature embedding
      $$ a^i_{opt} = f_{de}(h^{i, r}_{trans}) $$
      \State Reconstruct the one-hot image transformation vectors
      $$\tilde{c}^i_1, \tilde{c}^i_1, \cdots, \tilde{c}^i_M = FC(a^i_{opt})$$
      \State Achieve the corresponding image transformation operations $\tilde{t}^i_1, \tilde{t}^i_1, \cdots, \tilde{t}^i_M$
    \EndFor
    \State $$x^{adv}_0 = x, \quad g_0 = 0$$
    \For {i $\in \{0, \cdots, T-1\}$ }
      \State
      \begin{gather*}
      z = \frac{1}{N} \sum_{j=1}^N \frac{\nabla_{x_t^{adv}} \mathcal{J}(f(\tilde{t}^j_M \circ \cdots \circ \tilde{t}^j_1(x_t^{adv})), y)}{\|\nabla_{x_t^{adv}} \mathcal{J}(f(\tilde{t}^j_M \circ \cdots \circ \tilde{t}^j_1(x_t^{adv})), y)\|_1} \\
      g_{t+1} = \mu \cdot g_t + z \\
      x_{t+1}^{adv} = x^{adv}_t + \alpha \cdot sign(g_{t+1})
      \end{gather*}
    \EndFor
    \State \Return $x^{adv}_T$
  \end{algorithmic}
\end{algorithm}  

\end{document}